\documentclass[10pt,twocolumn,letterpaper]{article}

\usepackage[pagenumbers]{cvpr} 

\definecolor{cvprblue}{rgb}{0.21,0.49,0.74}
\usepackage[pagebackref,breaklinks,colorlinks,allcolors=cvprblue]{hyperref}
\usepackage{amsmath}
\usepackage{graphicx}
\usepackage{multicol}
\usepackage{multirow}
\usepackage{makecell}
\usepackage{listings}
\usepackage{minted}

\newcommand{\zhang}[1]{{\color{black} #1}}

\newcommand{\gz}[1]{{\color{black} #1}}

\newcommand{\wangcan}[1]{{\color{black} #1}}

\newcommand{\hjw}[1]{{\color{black} #1}}

\def\paperID{8235} 
\def\confName{CVPR}
\def\confYear{2026}

\title{DetAny4D: Detect Anything 4D Temporally in a Streaming RGB Video}

\author{
Jiawei Hou$^{1}$, Shenghao Zhang$^{2,*}$, Can Wang$^{3}$, Zheng Gu$^{4}$, Yonggen Ling$^{2}$, \\ Taiping Zeng$^{1}$, Xiangyang Xue$^{1,*}$, Jingbo Zhang$^{2,*}$
}

\begin{document}
\maketitle

\renewcommand{\thefootnote}{}
\footnotetext{
 $^{*}$ Co-corresponding author
}

\footnotetext{
 $^{1}$ Jiawei Hou and Xiangyang Xue are with School of Computer Science, Fudan University; Taiping Zeng is with Institute of Science and Technology for Brain-Inspired Intelligence, Fudan University. \textit{jwhou23@m.fudan.edu.cn} \textit{\{xyxue, zengtaiping\}@fudan.edu.cn}
}

\footnotetext{
 $^{2}$ Authors are with Tencent Robotics X, Tencent Binhai Building, Shenzhen, Guangdong 518057, China. \textit{\{popshzhang, eckertzhang, rolandling\}@tencent.com}
}

\footnotetext{
 $^{3}$ The University of Hong Kong. \textit{canwang@hku.hk}
}

\footnotetext{
 $^{4}$ Shenzhen University. \textit{guzheng@szu.edu.cn}
}

\begin{abstract}
Reliable 4D object detection, which refers to 3D object detection in streaming video, is crucial for perceiving and understanding the real world. Existing open-set 4D object detection methods typically make predictions on a frame-by-frame basis without modeling temporal consistency, or rely on complex multi-stage pipelines that are prone to error propagation across cascaded stages. Progress in this area has been hindered by the lack of large-scale datasets that capture continuous reliable 3D bounding box (b-box) annotations. 
To overcome these challenges, we first introduce DA4D, a large-scale 4D detection dataset containing over 280k sequences with high-quality b-box annotations collected under diverse conditions. Building on DA4D, we propose DetAny4D, an open-set end-to-end framework that predicts 3D b-boxes directly from sequential inputs. DetAny4D fuses multi-modal features from pre-trained foundational models and designs a geometry-aware spatiotemporal decoder to effectively capture both spatial and temporal dynamics. Furthermore, it adopts a multi-task learning architecture coupled with a dedicated training strategy to maintain global consistency across sequences of varying lengths. 
Extensive experiments show that DetAny4D achieves competitive detection accuracy and significantly improves temporal stability, effectively addressing long-standing issues of jitter and inconsistency in 4D object detection. Data and code will be released upon acceptance.

\end{abstract}    
\section{Introduction}
\label{sec:intro}

\begin{figure}[ht]
\centering
\includegraphics[width=\linewidth]{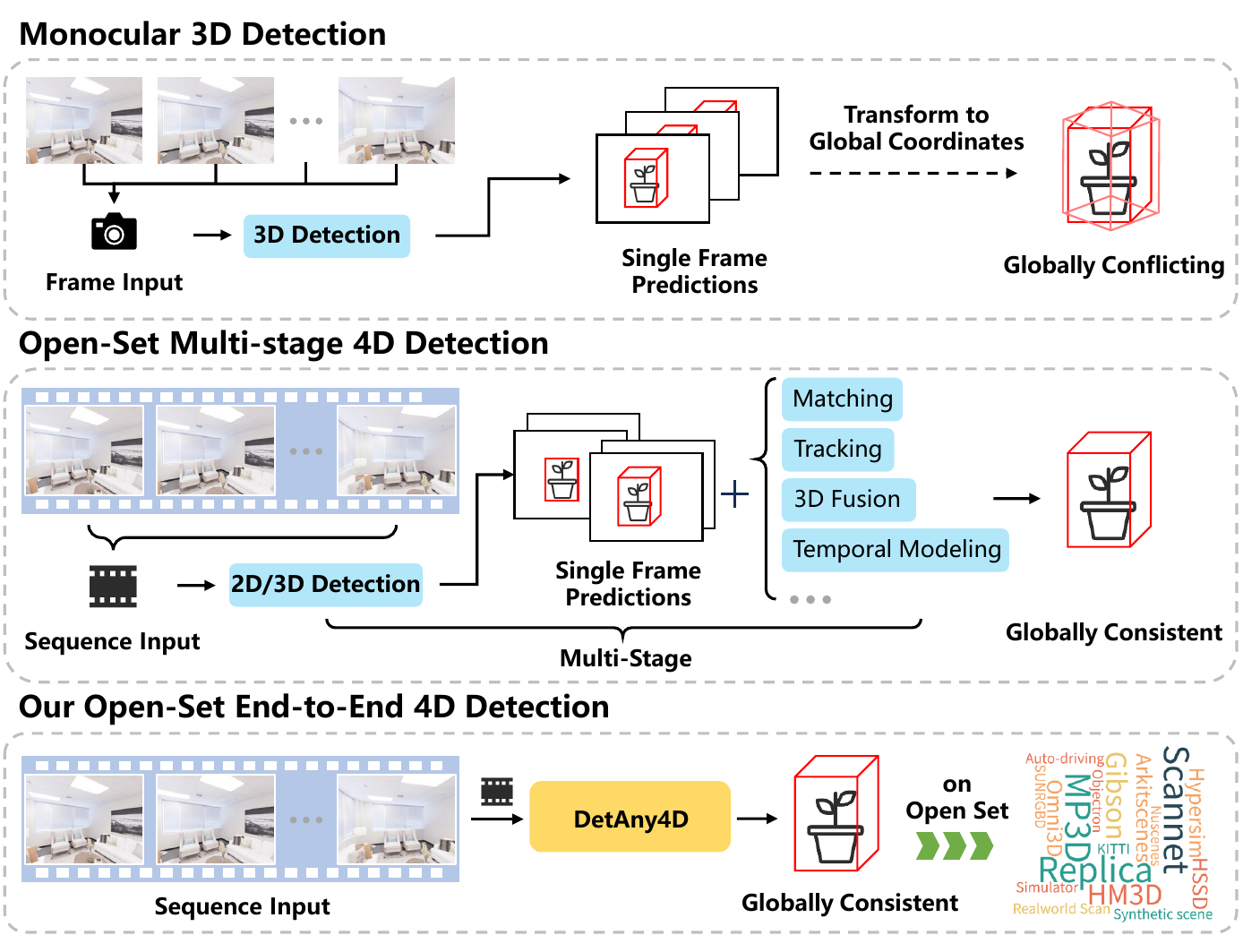}
\vspace{-0.3in}
\caption{\wangcan{Comparison with existing methods. Existing 3D detectors predict on a frame-by-frame basis, which causes inconsistency when transforming 3D b-boxes into global coordinates. 
Current open-set 4D detectors typically address 3D predictions and cross-frame relationships in a multi-stage manner, which is complex and prone to error propagation across cascaded stages. 
In contrast, we propose an open-set end-to-end 4D detection benchmark that directly predicts globally consistent 3D b-boxes.}}
\vspace{-0.2in}
\label{fig::teaser}
\end{figure}

\wangcan{
3D object detection serves as a pivotal task to perceive the world. 
}
\wangcan{
Previous methods have made significant strides in predicting the 3D bounding box (b-box) from single snapshots~\cite{rukhovich2023tr3d, zhang2025detect, brazil2023omni3d} and pre-scanned point clouds~\cite{mao2025spatiallm, openpcdet2020}.
}
\wangcan{However, when applied to streaming videos, these methods are either prone to error propagation and accumulation or depend on prior global environment scanning.} 
\wangcan{Since streaming 3D bounding box prediction is crucial for reliable long-term reasoning, predicting future states, and ensuring stable decision-making in real-world applications, we focus on this task and formulate it as 4D object detection, which integrates 3D information with temporal awareness.
}



\wangcan{
An intuitive solution is to directly apply the single-frame 3D detection methods~\cite{rukhovich2023tr3d, zhang2025detect, brazil2023omni3d} to 4D detection (Figure \ref{fig::teaser}-top), but this would lead to inconsistency because temporal relationships are not considered.
In response to this, another approach~\cite{gu2024conceptgraphs, tairos2025}  to 4D object detection is a multi-stage pipeline that post-processes single-frame detections with tracking and 3D association modules to enforce temporal coherence (Figure \ref{fig::teaser}-middle).
Although feasible, these approaches often suffer from cumbersome design and limited robustness, leading to fragile behavior and error accumulation.
Several autonomous-driving studies have extended 3D detection to the spatiotemporal setting by incorporating ego-motion estimation, velocity prediction, and multi-stage detectors with temporal-consistency learning~\cite{brazil2020kinematic, he20233d}. However, these methods are typically tailored to specific driving scenarios and cover only a limited range of object categories. Consequently, they either lack open-set detection capability or still depend on multi-stage pipelines that are susceptible to error propagation.
}


\wangcan{
Despite these explorations, 4D object detection remains an open challenge: 1) A large-scale, high-quality 4D dataset with 3D b-box annotations is still lacking, largely due to the prohibitive cost and complexity of spatiotemporal annotation; 2) In temporal modeling, it remains unresolved how to build mechanisms that maintain long-term memory yet still detect objects accurately in newly arriving frames; 3) It is essential to devise methods that handle arbitrarily long sequences and deliver consistent, stable 3D detections under dynamically changing viewpoints.
}


\wangcan{
We tackle the first challenge by introducing DA4D, a large-scale 4D object detection dataset. First, we collect large-scale RGB sequences with depth and camera poses by moving the camera in the Habitat simulator~\cite{szot2021habitat}. Next, we project object b-boxes in global coordinates to ego views and filter severely-occluded or out-of-view objects. Finally, we implement an adaptive strategy to adjust the 3D b-box annotations, taking into account factors such as visibility, occlusion, and historical observations, especially for objects with complex geometry. The DA4D dataset unifies 12 datasets of more than 280k sequences with spatiotemporal-aligned annotations.}


\wangcan{
Leveraging this large-scale dataset, we propose DetAny4D, an open-set end-to-end framework that addresses both remaining challenges, rather than relying on a complex multi-stage design (Figure \ref{fig::teaser}-bottom).
DetAny4D fuses visual features extracted by SAM~\cite{kirillov2023segment} and DINO~\cite{oquab2023dinov2}, which are then fed into the proposed Spatiotemporal Decoder, along with multi-modal prompt tokens and b-box tokens. 
DetAny4D further introduces a multi-head architecture for multi-task learning, enabling camera, depth, and 3D b-box estimations, while ensuring effective model training.
}


\wangcan{
To enable effective training on sequential data and handle variable-length sequences during inference, we propose a dedicated training strategy with random cropping and object query padding for newly appearing objects. 
Furthermore, a dedicated loss function is designed to enforce global consistency among the b-boxes across the sequence.}





\wangcan{Our contributions are summarized as follows:
\begin{itemize}
    \item We introduce DA4D, a large-scale 4D object detection dataset consisting of 280k sequences, each with spatiotemporal-aligned 3D b-box annotations, depth data, and camera pose information.
    \item We propose DetAny4D, an open-set end-to-end 4D detection benchmark which predicts spatiotemporally aligned 3D b-boxes across frames.
    \item We propose a SpatioTemporal Decoder module integrated with a multi-task head framework and a tailored sequence-based training strategy, enabling robust cross-frame detections.
    \item Extensive evaluations show that DetAny4D achieves competitive 3D detection performance and a 10–30\% reduction in cross-frame variance compared to single-frame detection solutions, while also exhibiting strong open-set detection capability, demonstrating its overall effectiveness on the 4D detection task.
\end{itemize}
}

\section{Related Works}
\label{sec:related}

\begin{figure*}[ht]
\centering
\includegraphics[width=\linewidth]{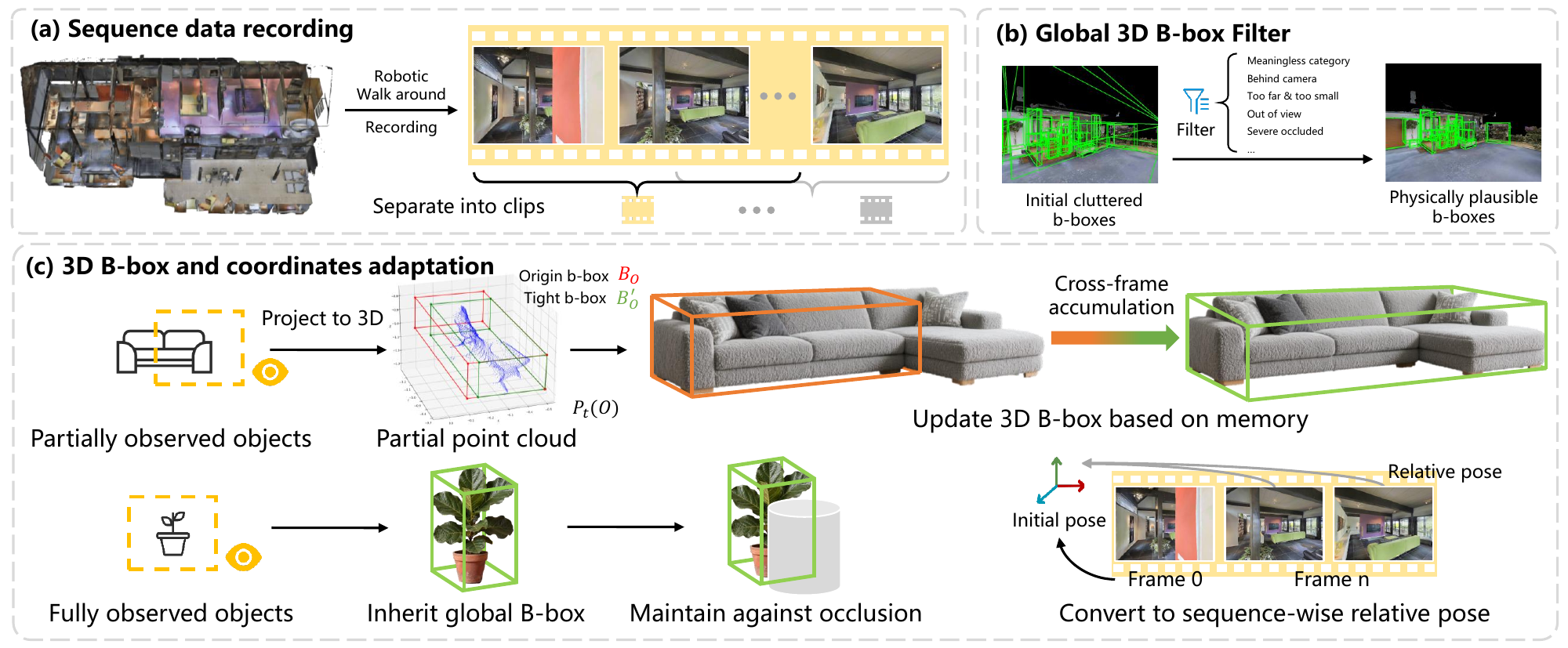}
\vspace{-0.3in}
\caption{The data processing pipeline for 4D detection task. We record posed RGB frames sequentially and separate the records into fixed-length sequences. Objects in global coordinates are projected into ego view and filtered with policies to delete occluded and out-of-view objects. Objects b-boxes are then recalculated according to the visibility and accumulated considering the point cloud within the sequence. Finally, the coordinates of a sequence is adapted referring to the first frame.}
\vspace{-0.2in}
\label{fig::data_pipeline}
\end{figure*}

\subsection{Monocular 3D Object Detection}
\label{sec:related:scene_understanding}

\zhang{Recently, advances in autonomous driving and embodied AI have made object detection a prominent focus in visual perception research. Early work primarily addressed the detection of specific categories of 3D objects within either indoor or outdoor scenes~\cite{chen2016monocular, dasgupta2016delay, huang2018cooperative, wang2021fcos3d, zhou2020iafa, liu2020smoke}. 
Cube R-CNN~\cite{brazil2023omni3d} utilizes the comprehensive Omni3D dataset to pioneer unified monocular 3D object detection, significantly improving generalization across diverse scene types. 
UniMODE~\cite{li2025towards} further advances the field by introducing a BEV-based monocular 3D object detector capable of seamless operation in both indoor and outdoor environments, thereby enhancing adaptability. 
OVMono3D~\cite{yao2024ovmono3d} innovates by decoupling 3D detection; it leverages a powerful 2D open-vocabulary detector for recognition, followed by a class-agnostic network for 3D lifting, enabling zero-shot detection of previously unseen categories. 
More recently, DetAny3D~\cite{zhang2025detect} proposes a foundation model for monocular 3D object detection by distilling knowledge from 2D foundation models.
Although such methods achieve satisfactory results on single frames, their lack of temporal context in continuous world perception often causes jitter and inconsistency. To address this, our approach focuses on video-based 3D detection, modeling spatiotemporal relationships to produce globally consistent 3D b-boxes.}


\subsection{Video Object Detection}
\label{sec:related:video_perception}

\zhang{To achieve continuous world perception, many existing works focus on predicting continuous 3D bounding boxes in video sequences~\cite{gu2024conceptgraphs, brazil2020kinematic, he20233d, rukhovich2023tr3d}. For example, Kinematic3D~\cite{brazil2020kinematic}, developed for autonomous driving, predicts 3D boxes frame by frame and subsequently applies ego-motion compensation and a Kalman filter to fuse detections across frames. BEVFormer~\cite{li2203bevformer} creates BEV representations by aggregating multi-camera features with a spatiotemporal transformer. TR3D~\cite{rukhovich2023tr3d} enables real-time 3D object detection for limited categories in indoor environments through sparse convolution and lightweight design. BADet~\cite{he20233d} adopts a two-stage process of initial detection and multi-frame fusion, incorporating global optimization of object centers and bundle adjustment of feature metrics to learn temporal dependencies. ConceptGraphs~\cite{gu2024conceptgraphs} and TAIROS~\cite{tairos2025} project 2D detections into 3D using depth, enabling category-wise fusion of object point clouds. However, these methods often rely on multi-stage pipelines or are restricted to specific scenarios, limiting their generalization. In contrast, our approach leverages open-vocabulary setting and an end-to-end framework for efficient, globally consistent 3D object detection in video sequences.}


\subsection{Point-Cloud-Based 3D Detection}
\zhang{To achieve globally consistent detections, another pipeline leverages complete 3D point clouds of an entire scene. Methods such as VoteNet~\cite{qi2019deep} use Hough voting for object center proposals, H3DNet~\cite{zhang2020h3dnet} integrates hybrid geometric primitives for greater accuracy, and RGBNet~\cite{wang2022rbgnet} models surface geometry with ray-based grouping. Multi-modal approaches like MTC-RCNN~\cite{park2021multi} fuse RGB images and point cloud features, while SpatialLM~\cite{mao2025spatiallm} employs large language models to convert point clouds to structured text, enabling interpretable and zero-shot detection. However, these methods require complete pre-scanned point clouds, resulting in high data and computational costs that limit scalability. In contrast, our approach produces consistent 3D bounding boxes directly from RGB sequences, achieving stronger generalizability without these constraints.}
\section{Data Generation Pipeline}
\label{sec:method:data_pipeline}

\begin{figure*}[ht]
\centering
\includegraphics[width=\linewidth]{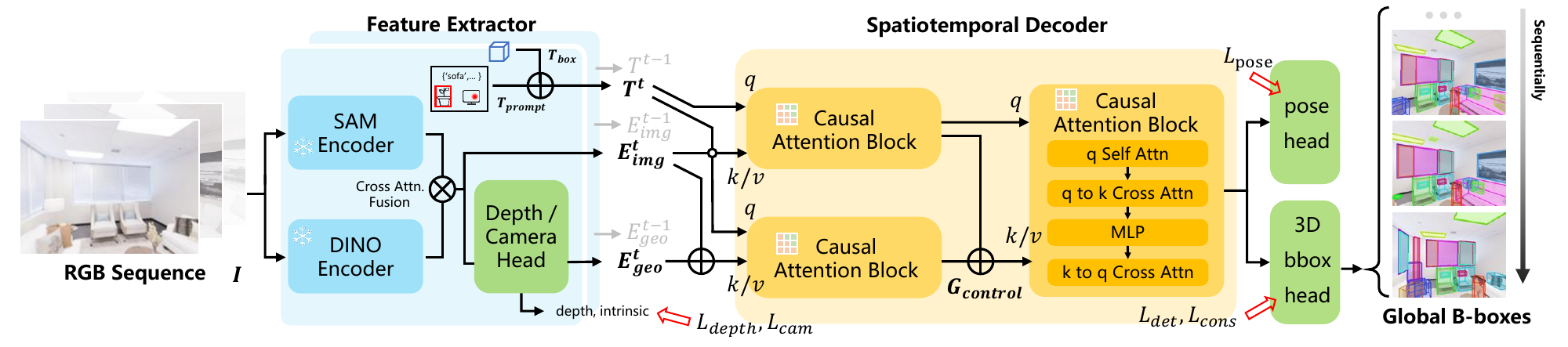}
\vspace{-0.3in}
\caption{Pipeline of the proposed DetAny4D model. RGB sequence with prompts are encoded with the feature extractor, generating tokens $T^t$, image embeddings $E_{img}^t$, and depth and camera-related embeddings $E_{d,m,c}^t$. A Geometry Context Transformer then inject 3D space embeddings in a transformer control manner, and together with embeddings decoded by the Spatiotemporal Transformer to generate prediction results. Multi-task heads are employed for effective training.}
\vspace{-0.2in}
\label{fig::model_pipeline}
\end{figure*}

We employ habitat simulator~\cite{szot2021habitat} to generate fine-labeled data for the 4D object detection task. As shown in Figure \ref{fig::data_pipeline}, the data processing pipeline consists of three main parts. 
\gz{First, we drive the robot to walk around in the virtual environment, record video sequences with an on-board camera, and divide them into fixed-lengh clips (Section \ref{sec:method:data_recording}). Meanwhile, Global coordinates object b-boxes are projected into ego frames and filtered according to designed strategy (Section \ref{sec:method:object_filter}). Then, the filtered b-boxes are adapted sequence-wise to get plausible ground-truth according to current view and memory. Coordinates are also adapted to relative coordinates in sequence (Section \ref{sec:method:box_adaptation}).}

\subsection{Sequence Data Recording}
\label{sec:method:data_recording}

As shown in Figure \ref{fig::data_pipeline}(a), we move an onboard camera through the scene in a random walk pattern, recording RGB observations and corresponding depth and poses. The recorded sequences are then segmented into fixed-length clips using a stride that is shorter than the clip length itself. This ensures an overlap between consecutive clips. Such design is intentional to facilitate the learning of object states under diverse initial visibility conditions within a sequence. For instance, it allows the model to learn from scenarios where an object may start fully within the field of view and gradually exit it, or conversely, begin outside the frame and progressively enter it.

\subsection{Global objects 3D B-box Filtering} 
\label{sec:method:object_filter}

\gz{To acquire globally consistent 3D object b-box as ground truth for the recorded clips, a natural idea it to project all objects from the world coordinates into the camera coordinates. However, such bare projection does not take severely occluded or out-of-view objects into consideration, which introduces significant artifacts (as shown in Figure \ref{fig::data_pipeline}(b)). Therefore, we apply a set of filtering rules to the initial cluttered b-boxes, keeping the physical plausibility while preserving the model's predictive capability on partially-occluded objects.}

\gz{We first filter out objects according to the semantic information. We remove background and non-object categories (e.g., floor, wall, void) in each frame.
Next, we exclude objects that are located behind the camera or far beyond a depth threshold.
Then, we filter out objects with severe occlusion. We calculate the projected 2D b-box in the ego view. For each 2D b-box, we count the semantic-map pixels belonging to the object. Objects with pixel counts below a threshold are removed, indicating strong occlusion.
We also project the eight vertices of each 3D b-box onto the current image plane and compare their depths with the depths of the corresponding 2D pixels. If more than five vertices lie behind their corresponding pixels, the object is considered heavily occluded and is filtered out.}

\begin{figure}[ht]
\centering
\includegraphics[width=\linewidth]{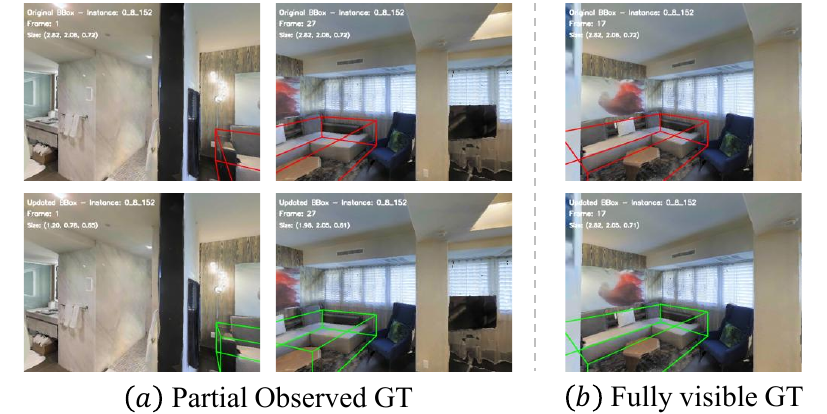}
\vspace{-0.25in}
\caption{Visualization of GT b-box adaptation strategy (Section \ref{sec:method:box_adaptation}). Taking an L-shaped sofa as example, red b-box denotes original GT and green denotes the adapted one. (a) shows adaptation of b-box when camera moves from left to right (column 1) and right to left (column 2) when sofa is partially observed. (b) shows when fully observed, adapted b-box aligns with original GT.}
\vspace{-0.2in}
\label{fig::box_adaptation}
\end{figure}

\subsection{3D B-box and Coordinates Adaptation} 
\label{sec:method:box_adaptation}


\gz{Unlike 3D detection, 4D detection exposes camera motion and changing viewpoints. As shown in Figure \ref{fig::data_pipeline}(c), considering an L-shaped sofa with a corner section, the initial global 3D b-box encompasses the entire sofa, including the corner. However, if the corner is not visible at the beginning, this global 3D b-box is intuitively unreasonable, and predicting it in early frames is physically implausible. 
A naive employment of such data leads to error propagation.
}

\gz{To further ensure physical credibility of the data, we propose a coordinate adaptation strategy to dynamically adjust the 3D b-boxes.}
If an object is fully visible at the start of a sequence, we maintain its original global 3D b-box until it completely disappears from the view or becomes permanently occluded. 
Conversely, if an object $O$ is partially visible at the beginning, we back-project the masked visible pixels into 3D space using their depth values. From this partial point cloud $P_t(O)$, we compute a tight-fitting b-box $B_O'$, which is smaller but shares the same orientation as the global 3D b-box. As the sequence progresses, we incrementally maintain and expand this point cloud as 
\begin{equation}
    P_t(O)=P_{t-1}(O)\cup \pi^{-1}(Mt(O),D_O),
\end{equation}
where $\pi^{-1}$ is the projection function. At the same time, we enlarge the computed b-box correspondingly. Once the object has been completely observe, we use the original global b-box as the ground truth and maintain it in remaining frames.
\gz{Such adaptation offers clear advantage for data construction. It enables our system to intelligently handle 3D objects with complex geometries, and also unlocks the model's capability to handle occlusions that evolve over time, which is crucial in temporal-aware 4D detection task.}
Figure \ref{fig::box_adaptation} visulize the ground truth adaptation (Section \ref{fig::data_pipeline}) that leverages a more reasonable accumulative b-boxes strategy.


Furthermore, regarding coordinate system handling, we define the camera pose of the first frame in a sequence as the initial reference pose. The camera poses for all subsequent frames are then calculated and stored as relative transformations with respect to this initial pose, and these relative poses are provided for the training process.

\hjw{Thus, processed by our pipeline, each sequence serves as an independent data instance. Its coordinate system is defined relative to the first frame, and its b-boxes are physically plausible for the current viewpoint. These boxes are updated incrementally based on the current observation and historical memory, which fully aligns with the objectives and logic of the 4D detection task.}

\section{Model Design}
\label{sec:method}

\hjw{As illustrate in Figure \ref{fig::model_pipeline}, DetAny4D takes an RGB sequence as input. We first employ a feature extractor to encode frame-wise inputs into image embeddings and introduce prompt and box tokens (Section \ref{sec:method:feature_extractor}). These embeddings are then injected into a unified 3D space and aggregated across frames using a spatiotemporal decoder (Section \ref{sec:method:decoder}).} \zhang{Multi-task heads are subsequently applied to supervise training and to produce temporally consistent 3D b-boxes for the queried objects throughout the sequence (Section \ref{sec:method:heads}). The overall training strategy and loss functions are specifically designed to address the challenges of sequence-level 4D object detection (Section~\ref{sec:method:train_strategy})}



\subsection{Feature Extractor}
\label{sec:method:feature_extractor}

\hjw{To extract semantic and geometry information from the input images, we follow \cite{zhang2025detect} and employ a feature extractor. Specifically, we adopt encoder modules from the pre-trained SAM~\cite{kirillov2023segment} and DINO~\cite{oquab2023dinov2} to capture priors from these 2D RGB sequence. These encoded features are further fused with a cross-attention-formed feature aggregator~\cite{zhang2025detect} to integrate multi-level embeddings $E_{img}^t$. To obtain geometric priors, we employ a depth and camera module, such as UniDepth-V2 \cite{piccinelli2025unidepthv2}, generating geometry-aware embedding $E_{geo}^t=\{E_{depth}^t, E_{cam}^t\}$, including depth embeddings $E_{depth}^t$ and camera parameter embeddings $E_{cam}^t$. At the same time, it predicts metric depth and camera parameters. Meanwhile, the prompts are encoded to tokens $T_{prompt}$ and concatenated with 3D b-box tokens $T_{box}$ to form the prompt embedding $T^t$, which serves as query input for subsequent transformer modules.}


\subsection{Spatiotemporal Decoder}
\label{sec:method:decoder}

\hjw{The spatiotemporal decoder accepts tokens and embbedings from sequence frames and generate spatiotemporal-aware implicit states for the following heads. The decoder compromises three Causal Attention Blocks (CAB).
Simultaneously, one CAB module integrates tokens $T^{1,\dots,t}$ and image embeddings $E_{img}^{1,\dots,t}$ and another module leverages tokens $T^{1,\dots,t}$ and geometry-aware embeddings $E_{geo}^{1,\dots,t}$, generating geometry control embedding $G_{control}$. Then a CAB module fuse the output from the former two modules and generating the final implicit features for following heads.}

\noindent\textbf{Causal Attention Block.} The causal attention block is a core component designed for modeling causal dependencies in sequential data, structured as a sequential stack of self-attention and bidirectional cross-attention, with causal masks integrated to strictly regulate information flow. \hjw{Causal mask, implemented as a lower triangular matrix sized according to the input sequence length and feature count, is applied to ensure information from current frame can be aware of historical information but blind to future values,
preventing information leakage.}
First, the self-attention module applies causal masking to $q$ to perceive the historical status.
Next, the bidirectional cross-attention guide causal-constrained information interaction between the tokens and embeddings. Overall, this design balances modeling flexibility and causal compliance with strict causal flow regulation, making it suitable for temporal-aware tasks.



\subsection{Multi-task Heads}
\label{sec:method:heads}

\zhang{To enable effective model training, we design a multi-head architecture for multi-task supervision. In the feature extractor stage, we employ the depth and camera head to estimate metric depth and camera intrinsics~\cite{piccinelli2025unidepthv2}. Additionally, following the spatiotemporal decoding stage, we introduce a camera pose head and a 3D b-box head, respectively. These heads infer the relative camera pose and 3D b-boxes for each frame based on the fused features. This design allows the model to account for camera view transformations and predict 3D b-boxes in global coordinates.}

\begin{figure*}[ht]
\centering
\includegraphics[width=\linewidth]{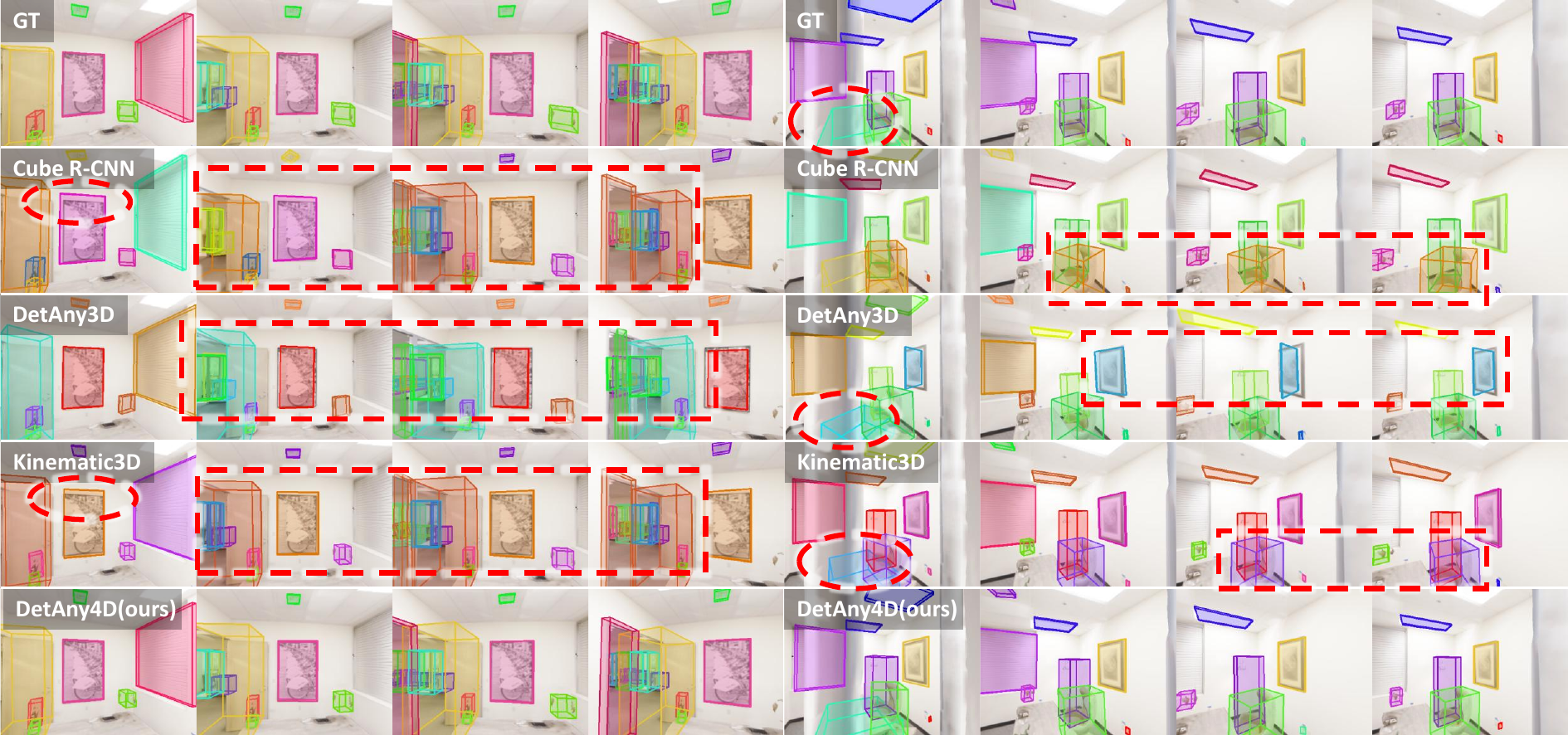}
\vspace{-0.25in}
\caption{Qualitative comparison with other methods on 3D b-box predictions across consecutive frames in a sequence. Our proposed DetAny4D predicts spatiotemporally aligned 3D b-boxes, while red circles and rectangles show inaccurate and inter-frame jtter predictions.}
\vspace{-0.2in}
\label{fig::main_experiment}
\end{figure*}

\subsection{Training}
\label{sec:method:train_strategy}


\noindent\textbf{Training Strategy.} \zhang{To enable effective training on sequential data and provide flexibility for variable-length sequences during inference, we introduce a dedicated training strategy. For input sequences, we randomly crop them within a fixed maximum sequence length, which promotes diversity in sequence lengths during training. To maintain the model’s ability for frame-wise detection, we also interleave single-frame 3D detection samples (represented as sequences of length one) into training batches.}
On the other hand, since the number of objects per frame is variable, we unify the number of prompts across all frames in a sequence before model input. We apply padding to the prompts to ensure the tokens from each frame have the same dimension in a sequence batch to be fed into the transformer. The loss function is calculated to disregard predictions associated with these padded prompts. During inference, the presence of these redundant tokens enables the model to detect newly appearing objects. \zhang{Further details can be found in the supplementary materials.}

\noindent\textbf{Loss Functions.} \zhang{To achieve spatiotemporal attention across input frames, we design a series of loss functions that balance the various capabilities of our multi-task heads.} we adopt widely used Scale-Invariant Logarithmic (SILog) loss~\cite{eigen2014depth, ummenhofer2017demon}, denoted as $L_{depth}$. Similarly, following the approach in UniDepth~\cite{piccinelli2025unidepthv2}, we supervise the camera intrinsics using the same SILog loss, denoted as $L_{cam}$.

The 3D b-box predictions are supervised with a combination of losses: 
\begin{equation}
    L_{det} = L_{center} + L_{d} + L_{IoU}^{2D} + L_{IoU}^{3D} + L_{corner} + L_{dim},
\end{equation}
\zhang{where $L_{center}$ and $L_{d}$ are both L1 losses, used to constrain the center position and depth of the 3D b-box, respectively. $L_{IoU}^{2D}$ and $L_{IoU}^{3D}$ represent the 2D and 3D IoU losses~\cite{openpcdet2020}, respectively, which are used to supervise the overall prediction quality. $L_{corner}$ denotes a permutation-invariant Chamfer Distance loss~\cite{brazil2023omni3d, yao2024open} applied to enforce rotation consistency. $L_{dim}$ is a soft loss designed to supervise the box dimensions. Unlike previous 3D detection methods that predict 3D b-boxes from a single image, our 4D detection objective aims to predict globally consistent 3D boxes across consecutive time frames. This temporal consistency makes it challenging to directly constrain the length, width, height, and rotation angle for geometrically symmetric objects observed from different viewpoints, since the definition of box dimensions can vary and may not be globally consistent across views. Therefore, for dimension supervision, the height is directly constrained using an L1 loss $L_{h}$. For the length and width, we alternately compute the differences between the predictions and ground truth, then apply a softmin operation to form a robust loss term. The overall dimension loss is thus defined as:}
\begin{equation}
    L_{dim} = L_h + \sum_{k=1,2} w_k l_{wl}^{(k)},
\end{equation}
\begin{equation}
    w_k = \frac{\text{exp}(-l_{wl}^{(k)}/\tau)}{\sum_{m=1}^2 \text{exp}(-l_{wl}^{(m)}/\tau))}, \tau = 0.1
\end{equation}
where
\begin{equation}
\begin{split}
    &\Pi_1 = 
    \begin{bmatrix}
    1 \quad 0 \\
    0 \quad 1
    \end{bmatrix},
    \Pi_2 = 
    \begin{bmatrix}
    0 \quad 1 \\
    1\quad 0
    \end{bmatrix} \\
    l_{wl} = &\Pi_1 \cdot (\text{L1}(\text{w}_{pred}, \text{w}_{GT})+\text{L1}(\text{l}_{pred},\text{l}_{GT})) + \\
    &\Pi_2 \cdot (\text{L1}(\text{w}_{pred}, \text{l}_{GT})+\text{L1}(\text{l}_{pred},\text{w}_{GT}))
\end{split}
\end{equation}

For the camera pose, following StreamVGGT~\cite{zhuo2025streaming}, we apply L1 loss $L_{pose}$ separately to the translation, rotation, and field of view components. \zhang{Furthermore, to address the sequential nature of the task, we design a consistency loss $L_{cons} = L_{spatial}+L_{temp}$ to enforce spatiotemporal consistency of the model’s predictions in the global coordinate system across different time frames. Here, $L_{spatial}$ is a Chamfer Distance loss computed between the transformed 3D boxes and the ground truth in the global world coordinates, while $L_{temp}$ is another Chamfer Distance loss that measures the discrepancy between per-frame predictions and their temporal average $\bar{B}_{w}$ across the sequence. Formally, these losses are defined as:
\begin{equation}
    L_{spatial} = \sum_i \text{chamfer}(B_{w}^i, B_{w}^{GT}),
\end{equation}
\begin{equation}
    L_{temp} = \frac{1}{T} \sum_i^T \text{chamfer}(B_{w}^i, \bar{B}_{w}),
\end{equation}
where $B_{w}^i = T_{c2w}^i \cdot B_{c}^i$ represents the 3D boxes transformed into the global coordinate system using the predicted box $B_{c}^i$ and camera pose $T_{c2w}^i$ for each frame in the sequence. $\bar{B}_{w}$ denotes the temporal average of these transformed 3D boxes. By leveraging these two losses, our model effectively ensures the consistency of its predictions in both spatial and temporal dimensions.
}

The total loss function is composed as follows:
\begin{equation}
    L = L_{depth}+L_{cam}+L_{det}+L_{pose}+L_{cons}.
\end{equation}


\noindent\textbf{Implementation Details.} We employ ViT-L DINOv2~\cite{oquab2023dinov2} and ViT-H SAM~\cite{kirillov2023segment} as image encoders. We employ pre-trained prompt encoder and 2D feature aggregator from DetAny3D~\cite{zhang2025detect}. The encoders and feature aggregator are frozen during training. 
The model is trained for 200 epochs taking 2 weeks. We leverage AdamW optimizer with initial learning rate of 0.0001, following cosine annealing policy. The image input is resized and padded to the resolution of 448. The maximum sequence length in training is set to 10.
\section{Experiments}
\label{sec:experiments}

\begin{table*}[tp]
\footnotesize
\centering
\setlength{\abovecaptionskip}{0.1cm}
\renewcommand{\arraystretch}{1.2}
\setlength{\tabcolsep}{4.5pt}
\begin{tabular}{ccccccccccccccccc}
\toprule 
\multirow{2}{*}{\textbf{Methods}} & \multirow{2}{*}{\textbf{Approach}} & \multicolumn{3}{c}{\textbf{Replica}} & & \multicolumn{3}{c}{\textbf{MP3D}} & & \multicolumn{3}{c}{\textbf{HM3D}} & & \multicolumn{3}{c}{\textbf{Full DA4D}} \\ \cline{3-5} \cline{7-9} \cline{11-13} \cline{15-17}
                    & & $\text{AP}_{\text{3D}}^\uparrow$ & $\text{Var}_{\text{v}}^\downarrow$ & $\text{Var}_{\text{c}}^\downarrow$ & & $\text{AP}_{\text{3D}}^\uparrow$ & $\text{Var}_{\text{v}}^\downarrow$ & $\text{Var}_{\text{c}}^\downarrow$ & & $\text{AP}_{\text{3D}}^\uparrow$ & $\text{Var}_{\text{v}}^\downarrow$ & $\text{Var}_{\text{c}}^\downarrow$ & & $\text{AP}_{\text{3D}}^\uparrow$ & $\text{Var}_{\text{v}}^\downarrow$ & $\text{Var}_{\text{c}}^\downarrow$ \\ \midrule
ImVoxelNet~\cite{rukhovich2022imvoxelnet} & \multirow{4}{*}{Monocular 3D Detection}& 10.03 & 1.24 & 1.36 & & 12.99 & 1.80 & 1.79 &  & 11.32 & 1.33 & 1.28 & & 11.98 & 1.50 & 1.43 \\
Cube R-CNN~\cite{brazil2023omni3d}& & 24.49 & 0.82 & 0.86 & & 20.90 & 1.71 & 1.68 &  & 21.84 & 0.94 & 0.89 & & 21.76 & 1.27 & 1.20 \\
OV Mono3D~\cite{yao2024ovmono3d}& & 23.81 & 0.97 & 0.93 & & 22.79 & 1.66 & 1.59 & & 25.38 & 0.96 & 0.91 & & 24.39 & 1.26 & 1.23 \\
DetAny3D ~\cite{zhang2025detect}& & \underline{27.25} & 0.62 & 0.52 & & \textbf{25.14} & 1.28 & 1.23 & & \textbf{26.80} & 0.83 & 0.70 & & \underline{27.16} & 0.99 & 0.90 \\ \midrule
Kinematic3D*~\cite{brazil2020kinematic} & \multirow{2}{*}{Video-Based 4D Detection} & 26.82 & \underline{0.58} & \underline{0.49} & & 24.83 & \underline{1.03} & \underline{0.97} & & 25.34 & \underline{0.74} & \underline{0.66} & & 25.46 & \underline{0.85} & \underline{0.78} \\
\textbf{DetAny4D(ours)} & & \textbf{28.02} & \textbf{0.56} & \textbf{0.47} & & \underline{24.98} & \textbf{0.90} & \textbf{0.86} & & \underline{26.79} & \textbf{0.62} & \textbf{0.53} & & \textbf{27.48} & \textbf{0.70} & \textbf{0.64} \\
\bottomrule 
\end{tabular}
\caption{4D detection evaluations on the DA4D dataset, comapred with 3D detection methods and multi-stage 4D detection methods. \textbf{Bold} and \underline{underlined} indicates the best and second-best results. Results on three sub-datasets and full DA4D are evaluated. $\text{AP}_{\text{3D}}$ reports Average Precision of predicted 3D b-boxes evaluated on each frame compared with GT b-boxes. $\text{Var}_{\text{v}}$ and $\text{Var}_{\text{c}}$ report the temporal variance of b-box centers and vertices belonging to the same instance among a sequence as introduced in Section \ref{sec:experiments:setups}. Kinematic3D* indicates realization with adapted rotation supervision for fairness (Section \ref{sec:experiments:setups}).}
\label{tab::main_res}
\end{table*}

\subsection{Setups}
\label{sec:experiments:setups}

\textbf{DA4D Benchmark.} 
The proposed DA4D benchmark is a unified 4D detection dataset that integrates 12 datasets covering tasks such as 4D detection, depth estimation, and reconstruction. It builds upon the Omni3D~\cite{brazil2023omni3d} dataset, which includes ArkitScenes~\cite{dehghan2021arkitscenes}, Hypersim~\cite{hypersim}, KITTI~\cite{Geiger2012kitti}, nuScenes~\cite{caesar2020nuscenes}, Objectron~\cite{objectron2021}, and SUNRGBD~\cite{song2015sun}, and adds sequential datasets including Replica~\cite{replica19arxiv}, MP3D~\cite{Matterport3D}, HM3D~\cite{ramakrishnan2021hm3d}, HSSD~\cite{khanna2023hssd}, Gibson~\cite{xiazamirhe2018gibsonenv}, and Scannet~\cite{dai2017scannet}, enriching resources for comprehensive 4D perception research.
Sequences are provided with depth maps, intrinsics, camera poses, prompts, and 3D b-boxes. The dataset includes 170k sequences with multiple frames for 4D task settings and 110k single frames to validate the single frame capacity. All licenses are considered according to policies.

\noindent\textbf{Baselines.} We compare following three types of methods:
\begin{itemize}
    \item \textbf{Monocular 3D Detection.} We employ DetAny3D~\cite{zhang2025detect}, OV-Mono3D~\cite{yao2024ovmono3d}, Cube R-CNN~\cite{brazil2023omni3d}, ImVoxelNet~\cite{rukhovich2022imvoxelnet} as single frame 3D detection baselines. DetAny3D~\cite{zhang2025detect} is a recently open-sourced open-set 3D detector, reaching state-of-the-art performance on hybrid datasets. We train baselines with our DA4D datasets in single frame manner and aligned training strategy.
    \item \textbf{Video-based 4D Detection.} We employ ConceptGraph*~\cite{gu2024conceptgraphs}, which projects frame-by-frame 2D detection results into 3D space using depth maps and performs object-level fusion to predict 3D bounding boxes. To ensure a fair comparison, we use depth predictions from \cite{piccinelli2025unidepthv2}, consistent with both \cite{zhang2025detect} and our own method. Kinematic3D*~\cite{brazil2020kinematic} is a close-vocabulary 4D detection approach specifically designed for autonomous driving. For fairness, we adapt its original rotation decomposition strategy to match the rotation supervision used in other baselines.
    \item \textbf{Point-Cloud-Based 3D Detection.} We conduct comparison with SpatialLM~\cite{mao2025spatiallm}, a structured scene understanding approach powered by large language models.
\end{itemize}

\noindent\textbf{Metrics.} We employ the widely used Average Precision (AP) for 3D b-boxes evaluation in each frame, using IoU3D thresholds range from $\tau \in [0.05, 0.10, \dots ,0.50]$. To evaluate the cross-frame b-box consistency, we transfer predicted 3D b-boxes of each object across frames $B_{1,\dots,t}$ into world coordinates $B'_{1,\dots,t}$ using the camera poses. We then compute the average b-box $\bar{B}$ for each object in the world coordinates. Subsequently, we calculate the Chamfer Distance between eight vertices and the center of the b-box in each frame and the average, denoted as 
\begin{equation}
    \begin{split}
        \text{Var}_{\text{v}}&= 1/t \sum \{\text{chamfer}(B_i', \bar{B})\}_{1,\dots,t}, \\ 
        \text{Var}_{\text{c}}&= 1/t \sum \{||\text{mean}(B_i')-\text{mean}(\bar{B})||_2\}_{1,\dots,t}.
    \end{split}
\end{equation}
Since not all methods predict camera poses, we employ the ground-truth camera poses for transformation to ensure a fair comparison. When compared with ConceptGraph*~\cite{gu2024conceptgraphs} and SpatialLM~\cite{mao2025spatiallm}, we employ F1 scores under IoU of 0.25 and 0.5 following \cite{mao2025spatiallm} instead of mAP, as suggested in \cite{avetisyan2024scenescript}. Because SpatialLM~\cite{mao2025spatiallm} does not produce confidence score and ConceptGraph*~\cite{gu2024conceptgraphs} maps predictions from various frames, which is hard to evaluate cross-frame confidence. Cross-frame variance is also not suitable to evaluate b-boxes predicted from point cloud.

\subsection{Main Results}
\label{sec:experiments:main_results}

As shown in Table \ref{tab::main_res}, we evaluate several approaches on the proposed DA4D dataset, including monocular 3D detection, video-based 4D detection, and our end-to-end 4D detection method. All methods are provided with the same globally-labeled 3D bounding boxes and trained with strategies kept as consistent as possible. The results demonstrate that our method achieves competitive 3D detection performance compared to state-of-the-art (SOTA) approaches on individual frames. Furthermore, in terms of inter-frame consistency, DetAny4D reduces temporal variance by 10\% to 30\% compared to leading single-frame 3D detectors and exhibits less jitter than previous video-based 4D detection methods. Figure \ref{fig::main_experiment} includes qualitative comparison with other methods, showing better temporal consistency of 3D b-boxes across consecutive frames. These findings confirm that our architectural design and training strategy effectively and efficiently enhance spatiotemporal consistency.


\subsection{Further Comparison}
\label{sec:experiments:zeroshot}

In Table \ref{tab::further_comparison}, we compare with SpatialLM~\cite{mao2025spatiallm} and ConceptGraphs~\cite{gu2024conceptgraphs}, two open-set global b-box prediction approaches. While our DetAny4D performs end-to-end 4D detection on RGB sequence, SpatialLM leverages pre-scanned global point-cloud for and ConceptGraphs relies on RGB-D sequence to perform multi-stage inference. SpatialLM works on the structured-scene assumption and predicts axis-aligned b-boxes, failing to perform well compared to the original GT. Results show that DetAny4D shows competitive performance against point-cloud based and multi-stage methods.

\begin{table}[tp]
\footnotesize
\centering
\setlength{\abovecaptionskip}{0.1cm}
\renewcommand{\arraystretch}{1.}
\setlength{\tabcolsep}{2pt}
\begin{tabular}{ccccc}
\toprule 
\textbf{Methods} & \textbf{Input} & \textbf{Strategy} & $\text{F1}_{\text{IoU@0.25}}^\uparrow$ & $\text{F1}_{\text{IoU@0.5}}^\uparrow$ \\ \midrule
SpatialLM~\cite{mao2025spatiallm} & \makecell{Pre-scanned \\ Point Cloud} & End-to-End & 27.1 & 12.8 \\ \midrule
ConceptGraph*~\cite{gu2024conceptgraphs} & RGB-D Seq. & Multi-Stage & 45.6 & 41.9 \\ \midrule
\textbf{DetAny4D(ours)} & RGB Seq. & End-to-End & \textbf{49.7} & \textbf{45.5} \\
\bottomrule 
\end{tabular}
\caption{4D detection comparison with point-cloud-based and multi-stage method. SpatialLM employs pre-scanned point cloud and can only predict axis-aligned b-box, failing to perform well compared with GT. ConceptGraph* utilizes predicted depth for fairness (Section \ref{sec:experiments:setups}), which relies on RGB-D sequence and multi-stage inference.}
\label{tab::further_comparison}
\end{table}

\begin{figure}[t]
\centering
\includegraphics[width=\linewidth]{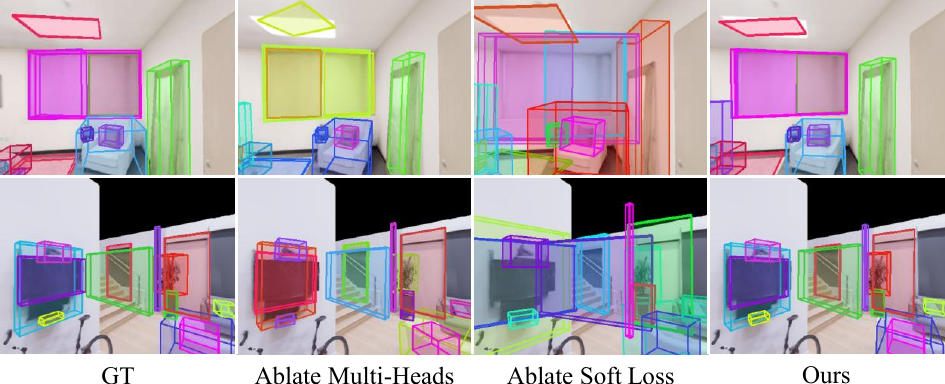}
\vspace{-0.15in}
\caption{Visualize effect of ablated designs. Columns from left to right show the GT, prediction from multi-heads-ablated realization, soft-loss-ablated realization, and our original approach.}
\vspace{-0.05in}
\label{fig::ablation}
\end{figure}

\subsection{Ablation Studies}
\label{sec:experiments:ablation}

\begin{table}[tp]
\footnotesize
\centering
\setlength{\abovecaptionskip}{0.1cm}
\renewcommand{\arraystretch}{1.2}
\setlength{\tabcolsep}{1.8pt}
\begin{tabular}{c|ccccc|ccc}
\toprule 
\textbf{Ablated} & Causal & Dim. & Vertices & \makecell{Depth \& \\ Cam.} & \makecell{Pose \& \\ Cons.} & $\text{AP}_{\text{3D}}^\uparrow$ & $\text{Var}_{\text{v}}^\downarrow$ & $\text{Var}_{\text{c}}^\downarrow$ \\ \midrule
Causal attn.                                & - & - & - & - & - & 26.78 & 0.95 & 1.01 \\ \midrule
\multirow{2}{*}{Soft loss}                  & $\surd$ & - & - & - & - & 26.84 & 0.93 & 0.98 \\
                                            & $\surd$ & $\surd$ & - & - & - & 26.88 & 0.91 & 0.96 \\ \midrule
\multirow{2}{*}{\makecell{Multi-\\heads}}   & $\surd$ & $\surd$ & $\surd$ & - & - & 27.15 & 0.60 & 0.51 \\
                                            & $\surd$ & $\surd$ & $\surd$ & $\surd$ & - & 27.28 & 0.59 & 0.50 \\ \midrule
\textbf{Ours}                               & $\surd$ & $\surd$ & $\surd$ & $\surd$ & $\surd$ & \textbf{27.29} & \textbf{0.54} & \textbf{0.48} \\
\bottomrule 
\end{tabular}
\caption{Ablation study of DetAny4D. Impact of each design on the $\text{AP}_{\text{3D}}$ and $\text{Var}_{\text{v}}$ is shown. Each component is progressively added to the origin model. The ablations are conducted on 10\% of the full training dataset to save computational cost.}
\vspace{-0.1in}
\label{tab::ablation}
\end{table}

In Table \ref{tab::ablation}, we ablate the key components of proposed DetAny4D, showing the performance degradation from DetAny4D to the original 3D detection model, which employs RGB sequence as input and performs single frame detection, ignoring temporal information. We also show qualitative comparison in Figure \ref{fig::ablation}, which shows that without the multi-heads design, the center and rotation pose of the 3D b-box converge poorly, while the absence of the soft loss strategy leads to significant errors in the dimensions.
\begin{itemize}
    \item \textbf{Causal attn.} By employing causal attention (Section \ref{sec:method:decoder}), the model is enabled to manage temporal information and perform sequence-wise training and inference. 
    \item \textbf{Soft loss.} The soft loss design described in \ref{sec:method:train_strategy} leads the model to learn the globally-annotated 3D b-boxes different from single frame 3D b-box annotation pattern. The order independent dimension and vertices chamfer distance loss make the training process a better convergence.
    \item \textbf{Multi-heads.} The multi-heads design described in Section \ref{sec:method:heads} equips the model with geometric and temporal transformation awareness. The depth and camera parameter prediction heads help mitigate gaps in multiple datasets training and provide dense supervision of the scene geometry. The camera pose head and consistency head model the view transformation and relative pose relationship of observed objects and camera view.
\end{itemize}

\section{Conclusion}
\label{sec:conclusion}

We propose DetAny4D, the first open-set end-to-end 4D detection benchmark, which predicts spatiotemporally aligned 3D b-boxes across frames. Our solution features a large-scale 4D object detection dataset with unified data generation pipeline, a novel framework with multi-task heads for coherent spatiotemporal reasoning, and tailored training strategy with hybrid loss functions. Extensive evaluations demonstrate that DetAny4D achieves competitive 3D detection accuracy with significantly temporal stability.
{
    \small
    \bibliographystyle{ieeenat_fullname}
    \bibliography{main}
}

\clearpage
\setcounter{page}{1}
\maketitlesupplementary

\section{Dataset Composition}
\label{sec:supp:dataset}

DA4D is a hybrid 4D detection dataset which includes 12 sub-datasets. The supporting tasks include sequence 4D detection, monocular 3D detection, depth estimation, and reconstruction tasks. DA4D is built on six datasets in Omni3D~\cite{brazil2023omni3d} (ArkitScenes~\cite{dehghan2021arkitscenes}, Hypersim~\cite{hypersim}, KITTI~\cite{Geiger2012kitti}, nuScenes~\cite{caesar2020nuscenes}, Objectron~\cite{objectron2021}, and SUNRGBD~\cite{song2015sun}) and expanded with a sequence dimension. Six more datasets are introduced for sequence-wise tasks, including Replica~\cite{replica19arxiv}, MP3D~\cite{Matterport3D}, HM3D~\cite{ramakrishnan2021hm3d}, HSSD~\cite{khanna2023hssd}, Gibson~\cite{xiazamirhe2018gibsonenv}, and Scannet~\cite{dai2017scannet}. The format is standardized similar to the Omni3D structure with additional sequence information. Each sequence compromises a list of frames. Each frame includes monocular RGB image, camera intrinsics, camera pose, depth map, and object information. The object information includes 3D b-box attribute ($[x,y,z,w,h,l,yaw]$), rotation pose, 2D b-box prompt, category, instance ID, and score.

\begin{figure}[h]
\centering
\includegraphics[width=\linewidth]{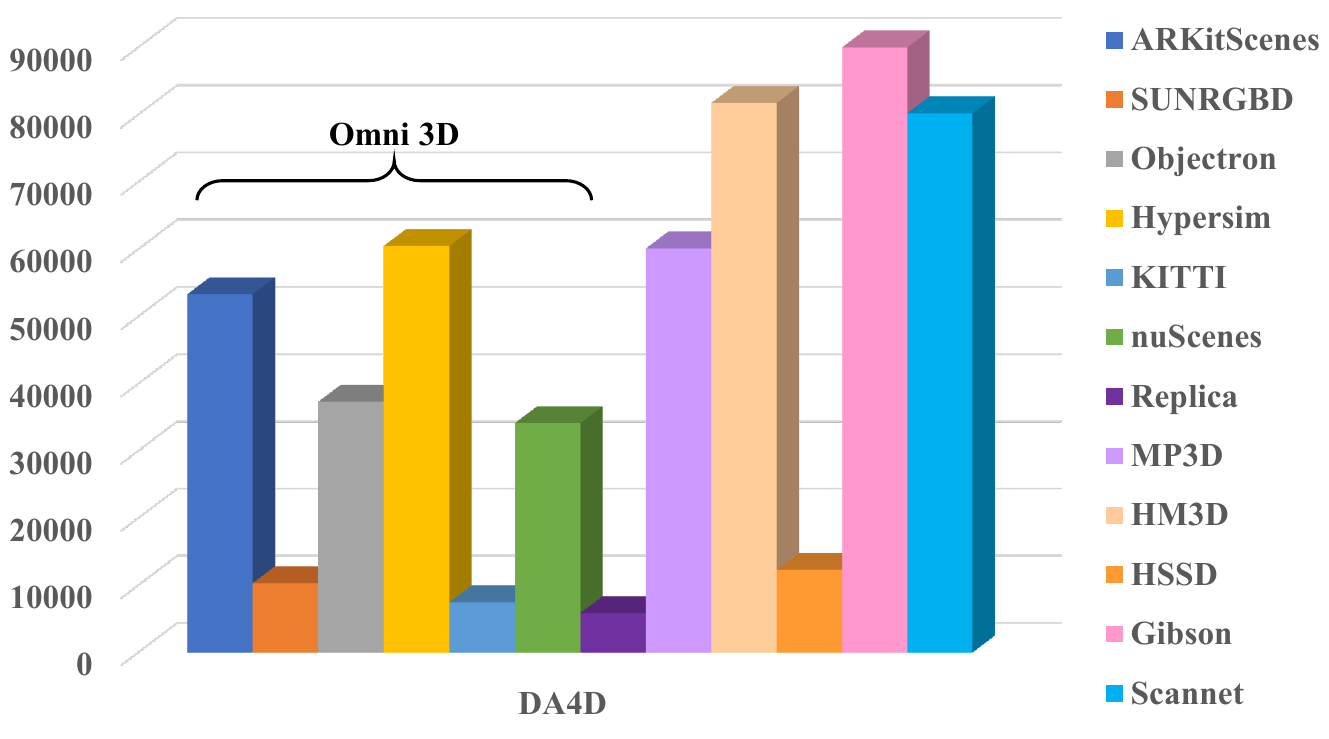}
\caption{Visualization of the DA4D dataset composition.}
\label{fig::sup:dataset}
\end{figure}

\textbf{Dataset Composition.} The dataset compromises original single frame 3D detection data as 3D detection capacity validation and multi-frame sequences data for 4D tasks. As shown in Figrue \ref{fig::sup:dataset}, DA4D consists of multi-frame sequences and Omni3D 3D detection data considered as sequences with length of 1.

\textbf{Dataset Split.} Sequences from Omni3D~\cite{brazil2023omni3d} follows the splitting strategy in prior works~\cite{brazil2023omni3d}. For newly compromised datasets to form multi-frame sequences, we split the scenes into the training and validation set, and separately record sequences from the selected scenarios, ensuring all scenes in the validation set have not been visited. 

\begin{figure}[h]
\centering
\includegraphics[width=\linewidth]{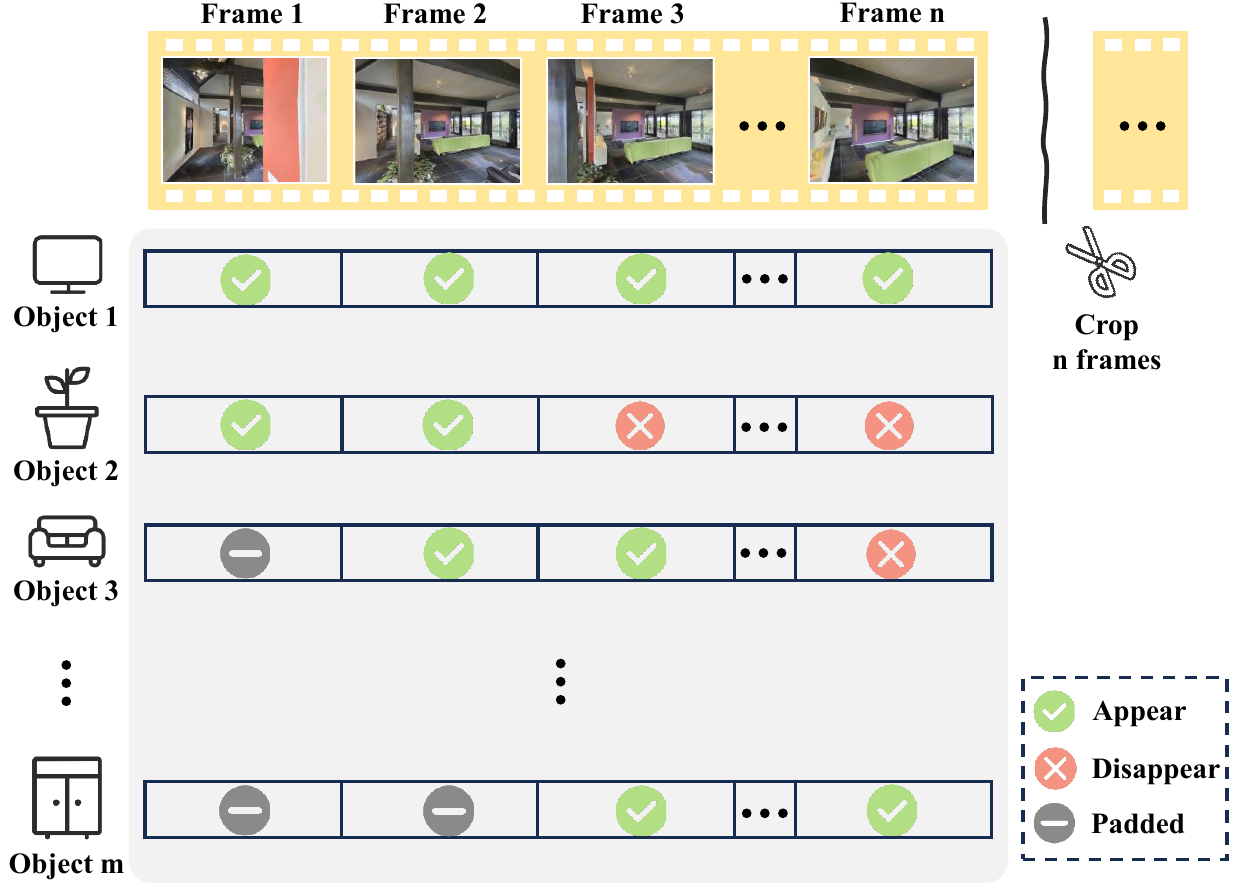}
\caption{Visualization of the sequence crop and object padding strategy. The object query list maintains the objects and padding status. As frames in the sequence forecast, object status updates. The disappeared objects and padded objects do not contribute to the loss.}
\label{fig::sup:padding}
\end{figure}

\section{Training Strategy Details}
\label{sec:supp:training}

Here we illustrate the sequence crop and object padding strategy in Section \ref{sec:method:train_strategy} with more details. As shown in Figure \ref{fig::sup:padding}, we crop each clip under a fixed maximum length. For the cropped sequence, we count the total objects number in each sequence and pad the objects in each frame to this number. During training, this strategy ensures each frame has the same object query dimension, and predictions generated with the padded queries will be masked which do not contribute to the loss. During inference, the padded queries enable the model to manage newly appeared objects during forecasting, where new objects are registered to the padded queries and maintained if the prediction result has a high score and differs against objects in the query memory list.

\begin{figure}[h]
\centering
\includegraphics[width=\linewidth]{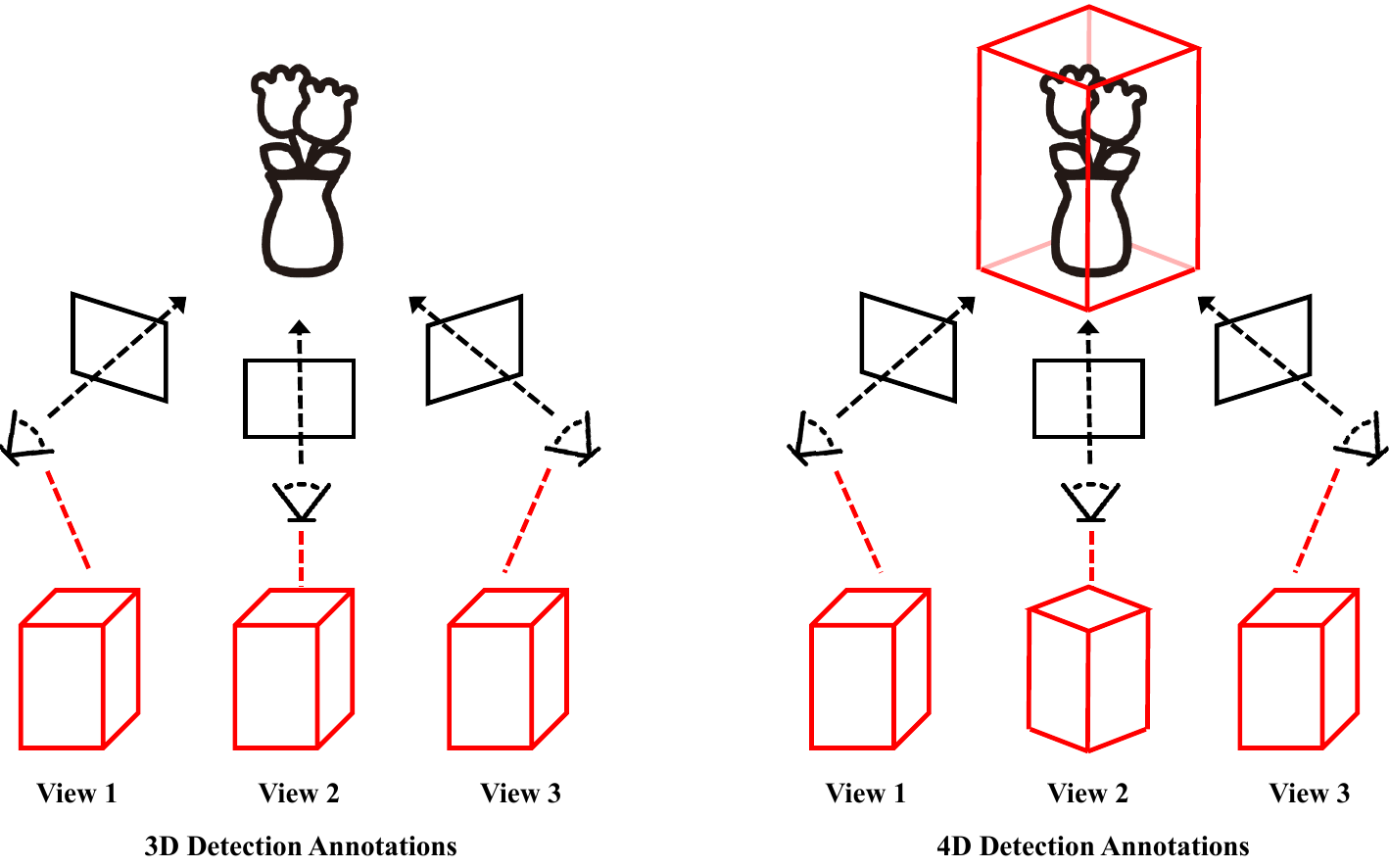}
\caption{Visualization of the annotation pattern differences. 3D detection task turns to annotate each object reffering only to the current view. 4D detection requires each view predicts objects in the global coordinates.}
\label{fig::sup:annotation}
\end{figure}

\section{3D B-box annotation Pattern}
\label{sec:supp:bbox}

This section illustrates the differences in the b-box annotation pattern for 3D detection task and 4D detection task. As shown in Figure \ref{fig::sup:annotation}, when observing an object from different views, 3D detection annotation pattern turns to consider each view as a singular observation and annotate the box referring only to the current view. On the contrary, for 4D detection, each object is registered with a b-box in the global coordinates and this constant b-box is projected to various views ensuring consistency globally. 

\begin{figure}[h]
\centering
\includegraphics[width=\linewidth]{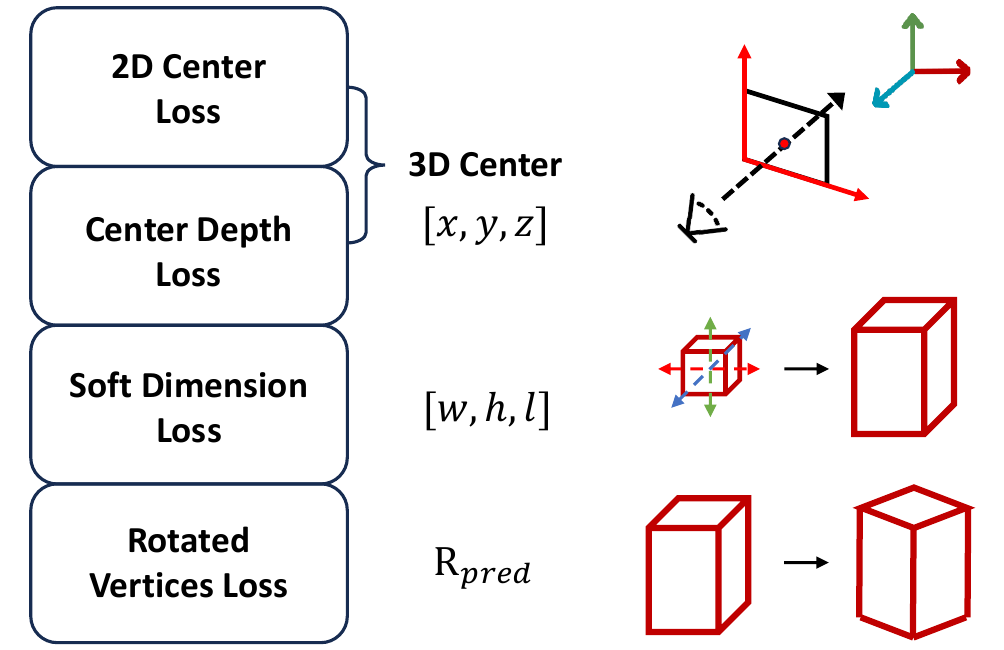}
\caption{Visualization of the b-box supervision. The 3D bounding box is constrained sequentially by its center, dimensions, and rotation angle.}
\label{fig::sup:loss}
\end{figure}

\begin{figure}[h]
\centering
\includegraphics[width=\linewidth]{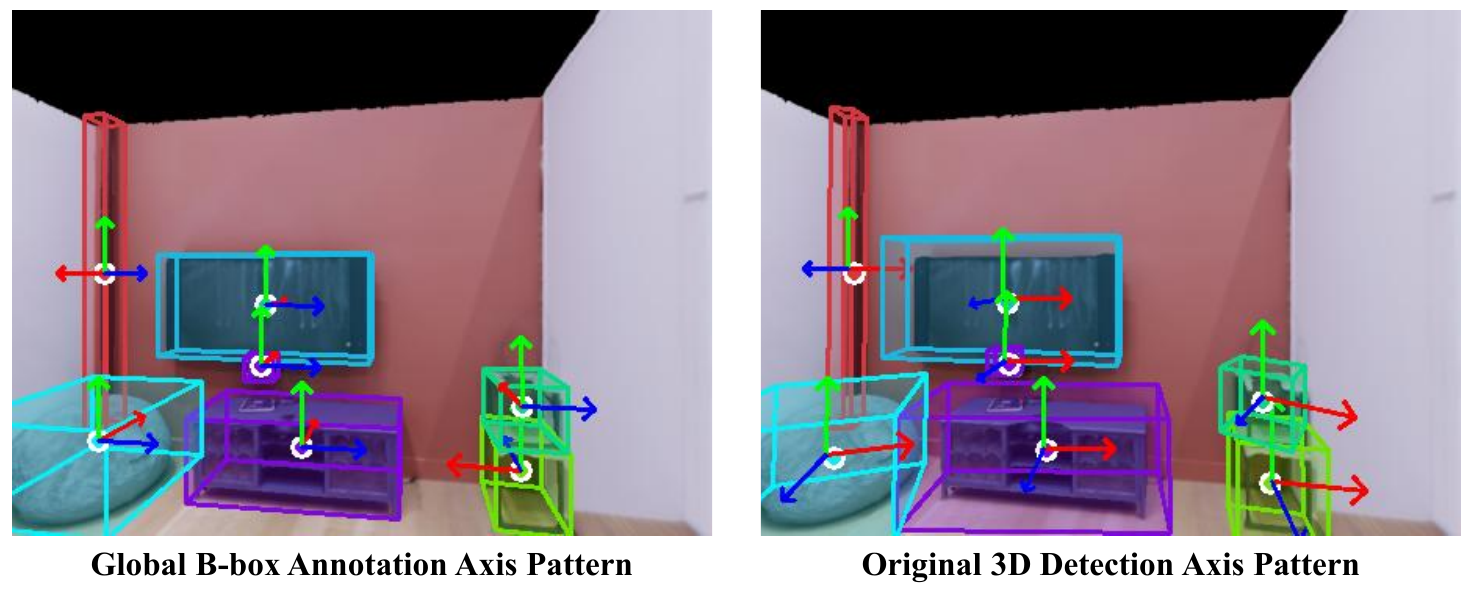}
\caption{Visualization of the axis pattern comparison between global annotations and 3D prediction results. Global annotated b-box dimensions and rotation differs from the predicted results of pre-trained 3D detection model.}
\label{fig::sup:axis}
\end{figure}

Given the annotation formats of the 4D detection task and our goal to leverage powerful prior knowledge from 3D detection, specifically by utilizing pre-trained 3D detection models, we designed a specialized loss function (Section \ref{sec:method:train_strategy}) to constrain the predictions to align with 4D task annotations. Our composite loss function, detailed in Section \ref{sec:method:train_strategy}, incorporates constraints on the center, dimensions, and rotation angle of the 3D bounding boxes. Figure \ref{fig::sup:loss} shows the supervision of the b-box.
In contrast to the rigid constraints traditionally used in 3D detection, we employ a softened loss for the dimensions and rotation. This design is motivated by the potential misalignment between the outputs of 3D pre-trained models and the 4D annotations, as shown in Figure \ref{fig::sup:axis}. Directly applying hard constraints can lead to slow and difficult convergence, whereas the softened loss facilitates a more effective and stable alignment of the predictions to the 4D ground truth.

\section{Open-Set Validation}
\label{sec:supp:openset}
To provide a detailed assessment of our model's open-set detection capabilities, we separately evaluated the 4D detection metrics on objects from categories that were within the training set (seen categories) and those that were outside of it (unseen categories) in the validation set. Table \ref{tab::sup:openset} shows the AP and F1 score under threshold IoU@0.5 (the same metrics illustrated in Section \ref{sec:experiments:setups}) of open-set categories and none-open-set categories in the validation set. It can be seen that the model performance on open-set highly keeps align with the none-open-set performance.

\begin{table}[tp]
\footnotesize
\centering
\setlength{\abovecaptionskip}{0.1cm}
\renewcommand{\arraystretch}{1.2}
\setlength{\tabcolsep}{2.2pt}
\begin{tabular}{cccccccc}
\toprule 
\multirow{2}{*}{\textbf{Datasets}} & \multirow{2}{*}{\makecell{Categories \\ Open-Set / All}} & \multirow{2}{*}{\makecell{Objects Num \\ Open-Set / All}} & \multicolumn{2}{c}{\textbf{Open-Set}} & & \multicolumn{2}{c}{\textbf{None-Open-Set}} \\ \cline{4-5} \cline{7-8}
    & & & $\text{AP}_{\text{3D}}^\uparrow$ & $\text{F1}^\uparrow$ & & $\text{AP}_{\text{3D}}^\uparrow$ & $\text{F1}^\uparrow$ \\ \midrule
Replica & 18 / 62 & 0.3k / 16k & 27.9 & 46.8 & & 28.0 & 47.1 \\
MP3D    & 20 / 330 & 3k / 98k & 24.7 & 41.9 & & 25.0 & 43.2 \\
HM3D    & 75 / 698 & 11k / 344k & 27.2 & 44.4 & & 26.7 & 43.7 \\
\bottomrule 
\end{tabular}
\caption{Evaluation of the open-set performance. We separately evaluate the AP and F1 score on three sub-datasets and also report the ratio of the categories and numbers of open-set objects.}
\label{tab::sup:openset}
\end{table}


\section{More Results}
\label{sec:supp:more_results}
We provide more visualization comparison to show the prior performance of our proposed DetAny4D on the spatiotemporal consistency validation.

\begin{figure*}[ht]
\centering
\includegraphics[width=\linewidth]{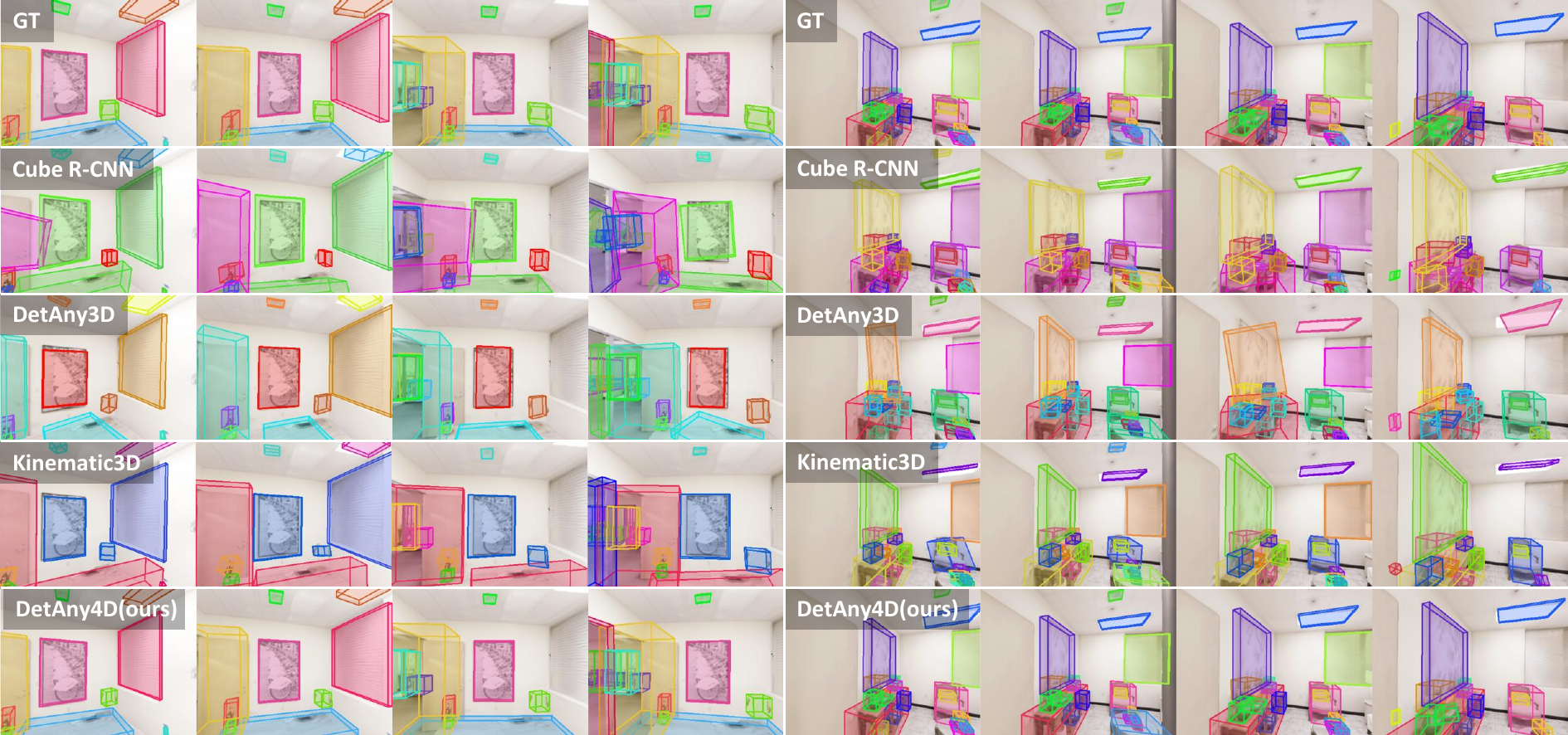}
\caption{More qualitative comparison results.}
\label{fig::exp1}
\end{figure*}

\begin{figure*}[ht]
\centering
\includegraphics[width=\linewidth]{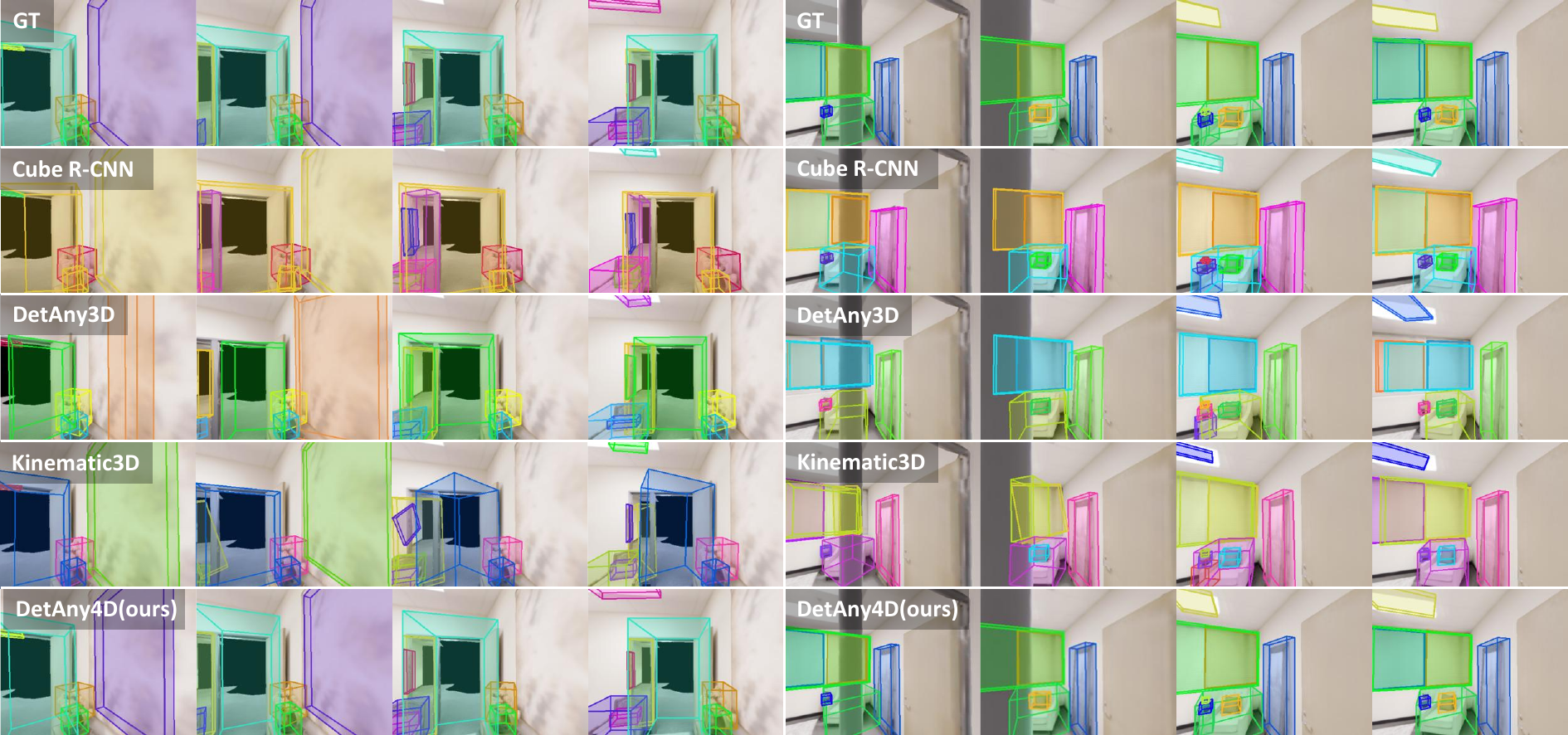}
\caption{More qualitative comparison results.}
\label{fig::exp2}
\end{figure*}

\begin{figure*}[ht]
\centering
\includegraphics[width=\linewidth]{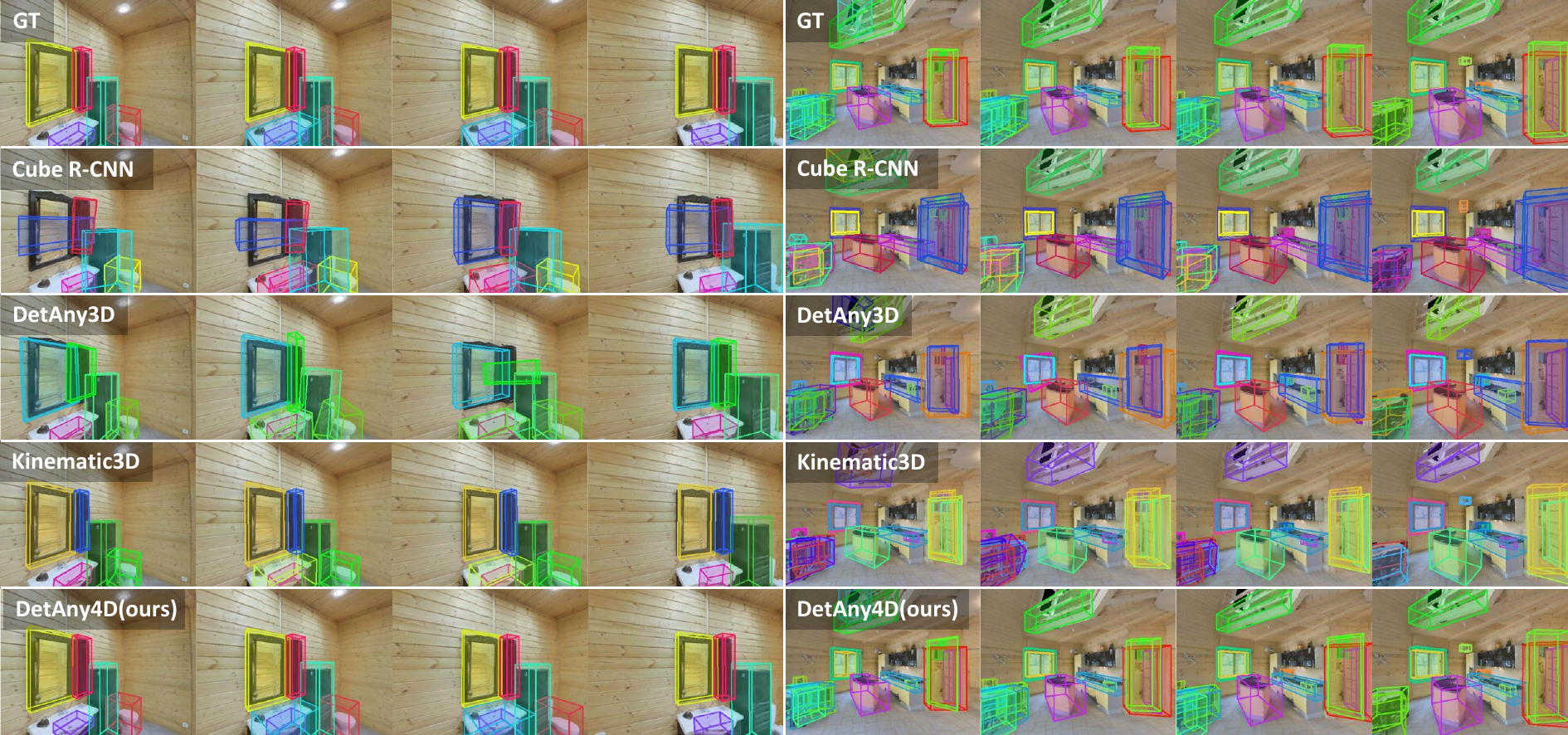}
\caption{More qualitative comparison results.}
\label{fig::exp7}
\end{figure*}

\begin{figure*}[ht]
\centering
\includegraphics[width=\linewidth]{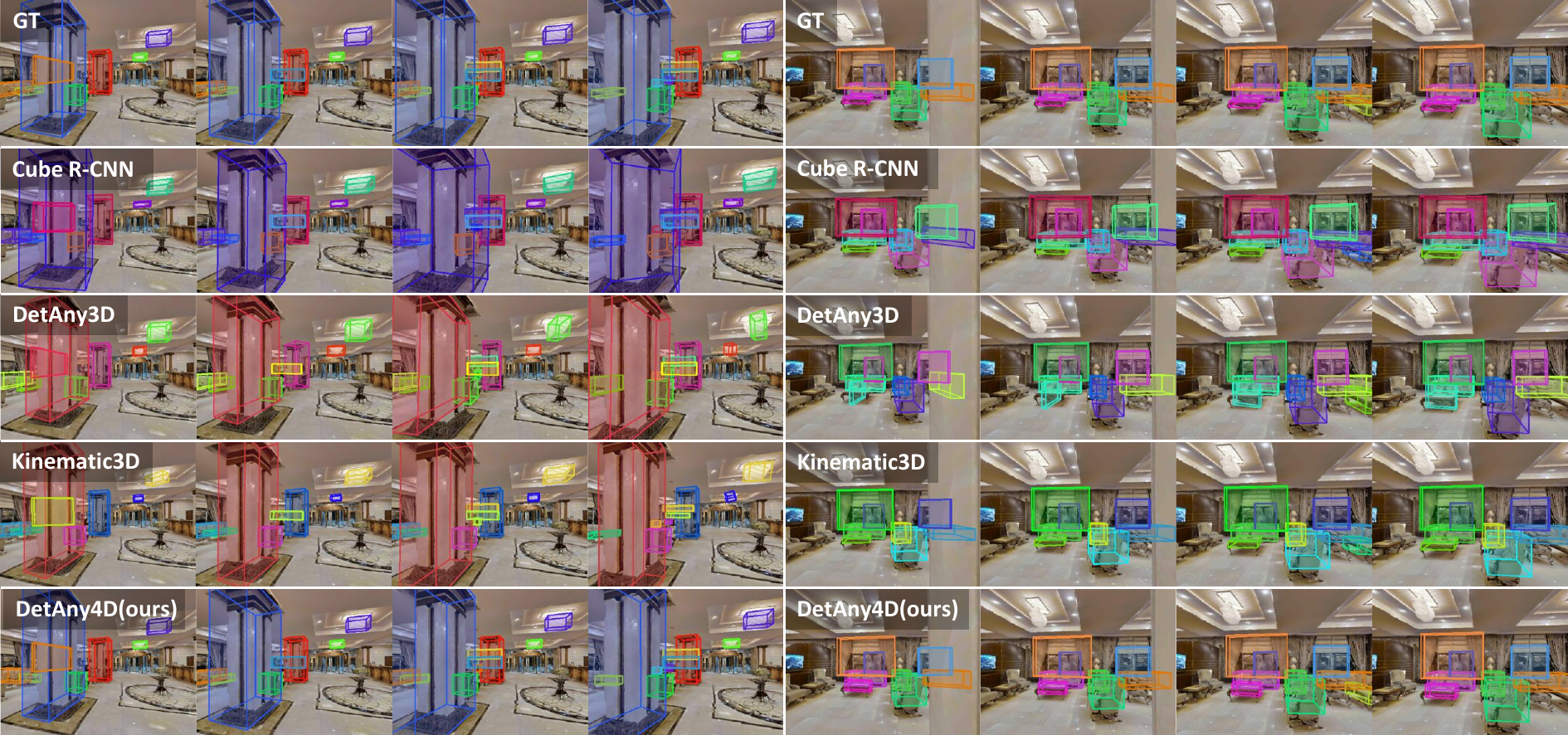}
\caption{More qualitative comparison results.}
\label{fig::exp8}
\end{figure*}

\begin{figure*}[ht]
\centering
\includegraphics[width=\linewidth]{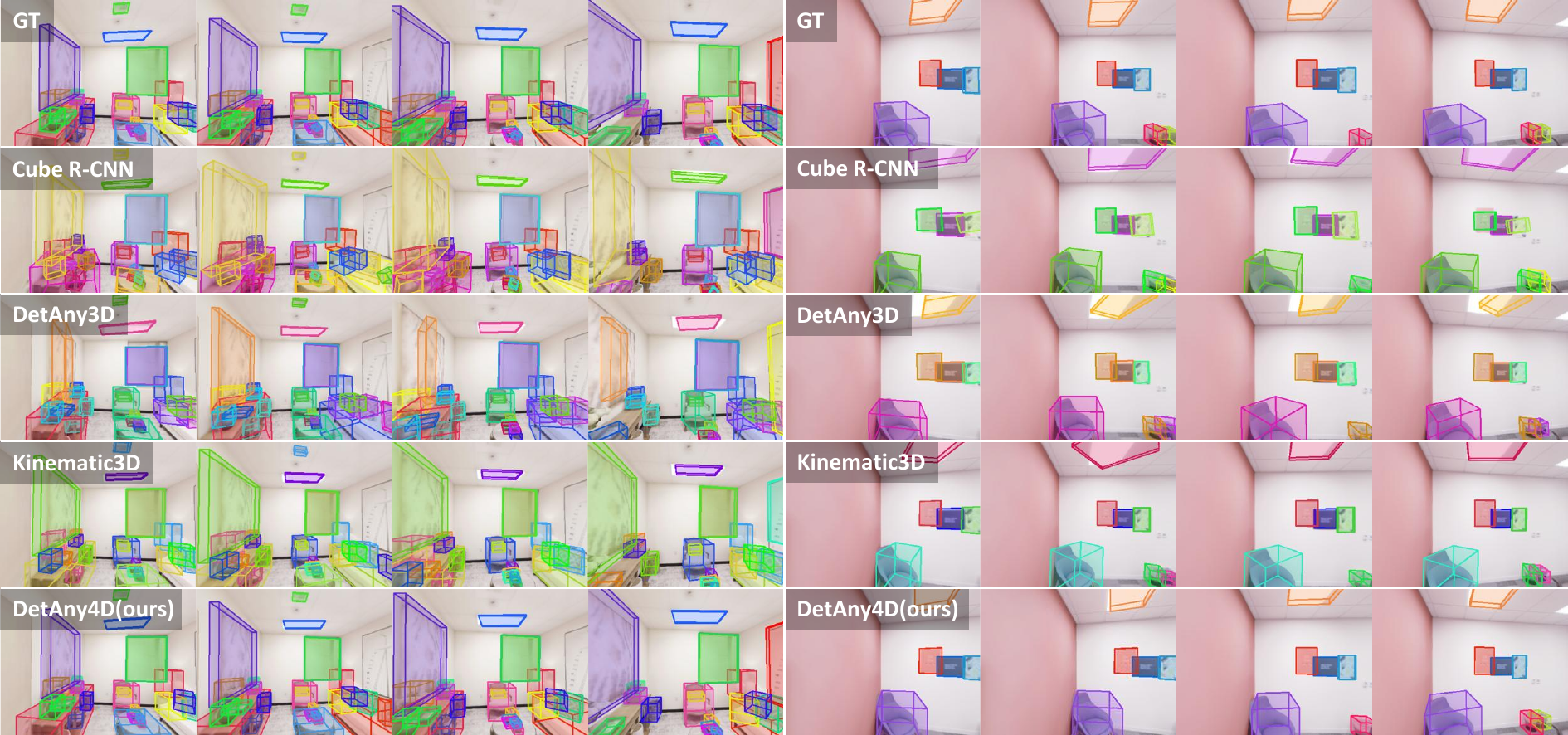}
\caption{More qualitative comparison results.}
\label{fig::exp3}
\end{figure*}

\begin{figure*}[ht]
\centering
\includegraphics[width=\linewidth]{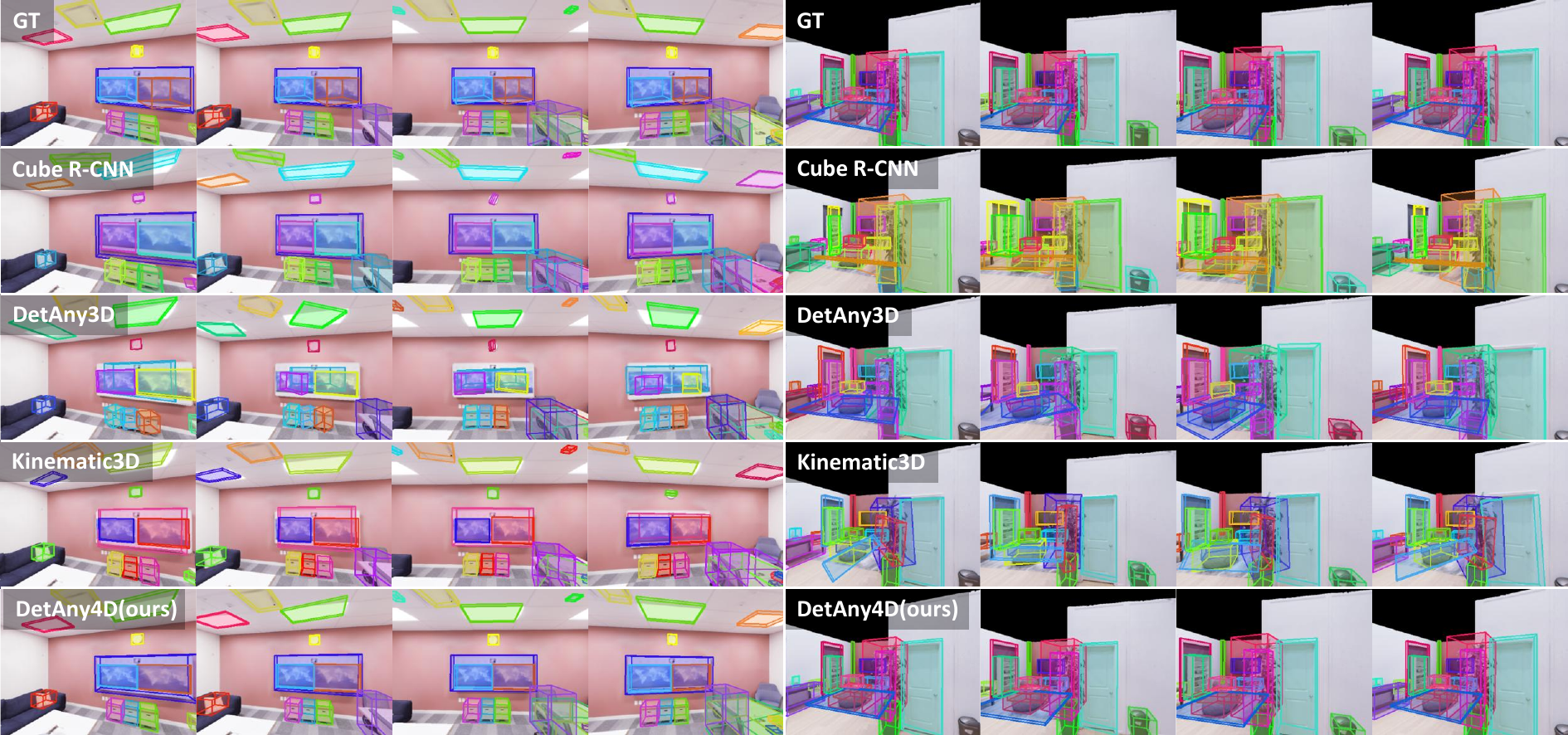}
\caption{More qualitative comparison results.}
\label{fig::exp4}
\end{figure*}

\begin{figure*}[ht]
\centering
\includegraphics[width=\linewidth]{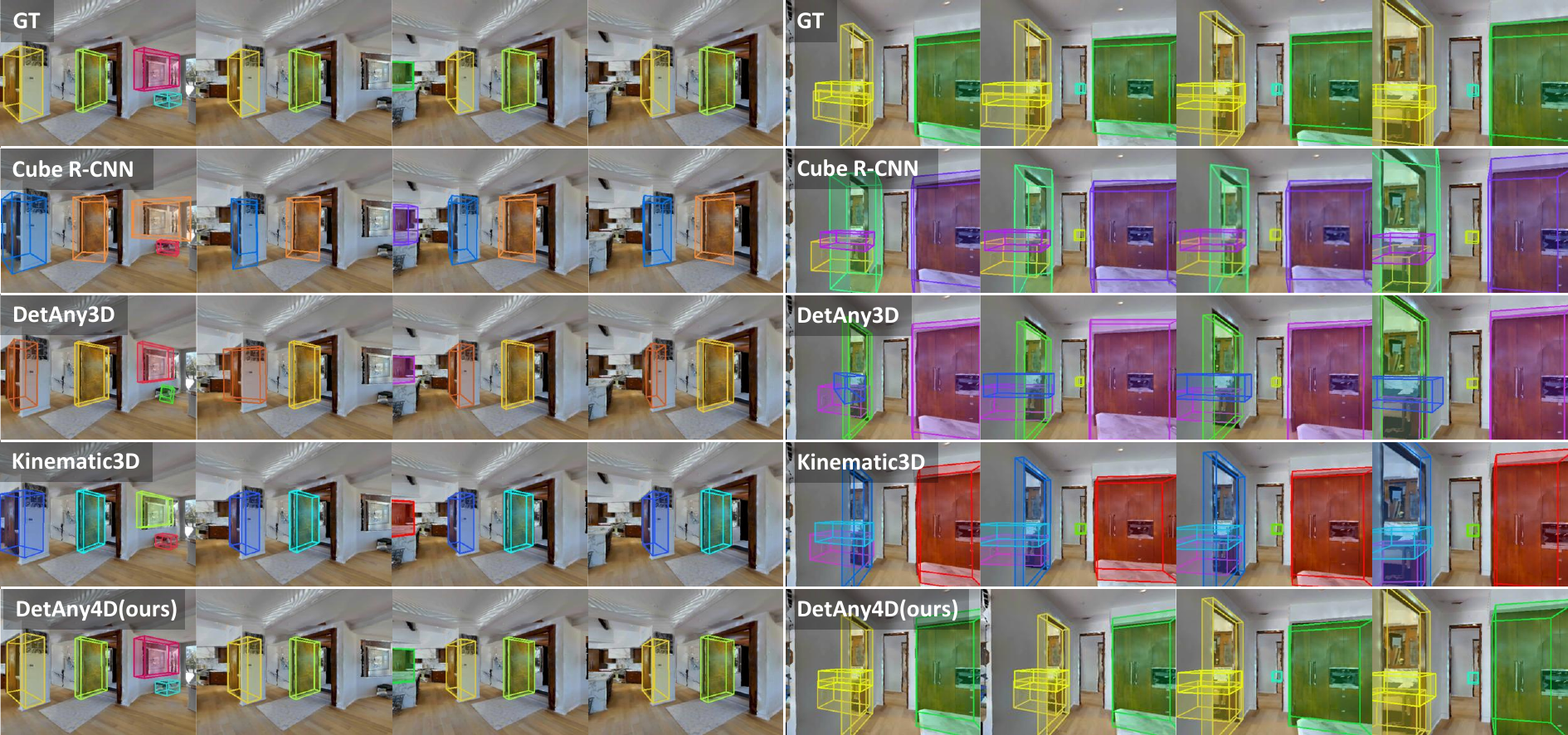}
\caption{More qualitative comparison results.}
\label{fig::exp9}
\end{figure*}

\begin{figure*}[ht]
\centering
\includegraphics[width=\linewidth]{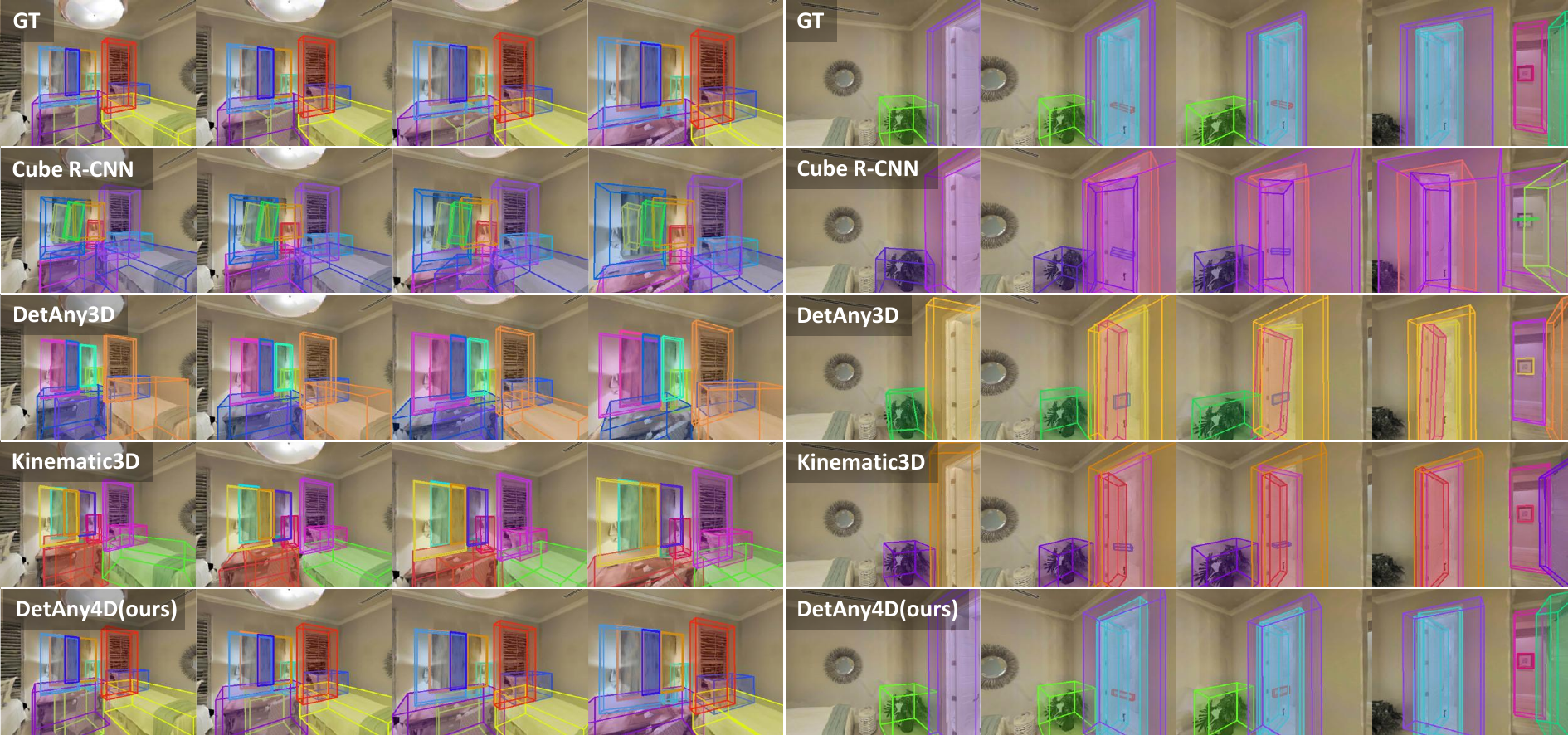}
\caption{More qualitative comparison results.}
\label{fig::exp10}
\end{figure*}

\begin{figure*}[ht]
\centering
\includegraphics[width=\linewidth]{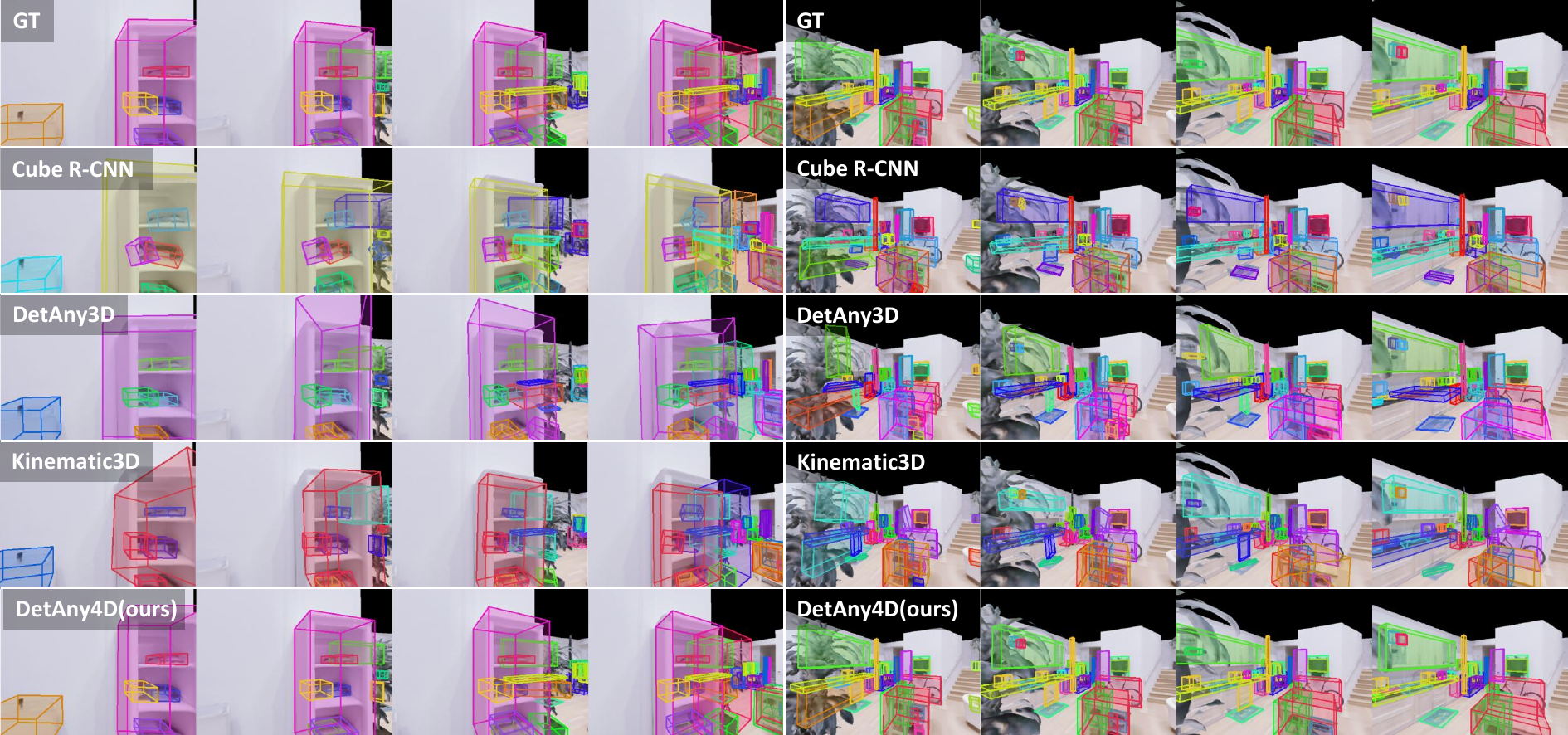}
\caption{More qualitative comparison results.}
\label{fig::exp5}
\end{figure*}

\begin{figure*}[ht]
\centering
\includegraphics[width=\linewidth]{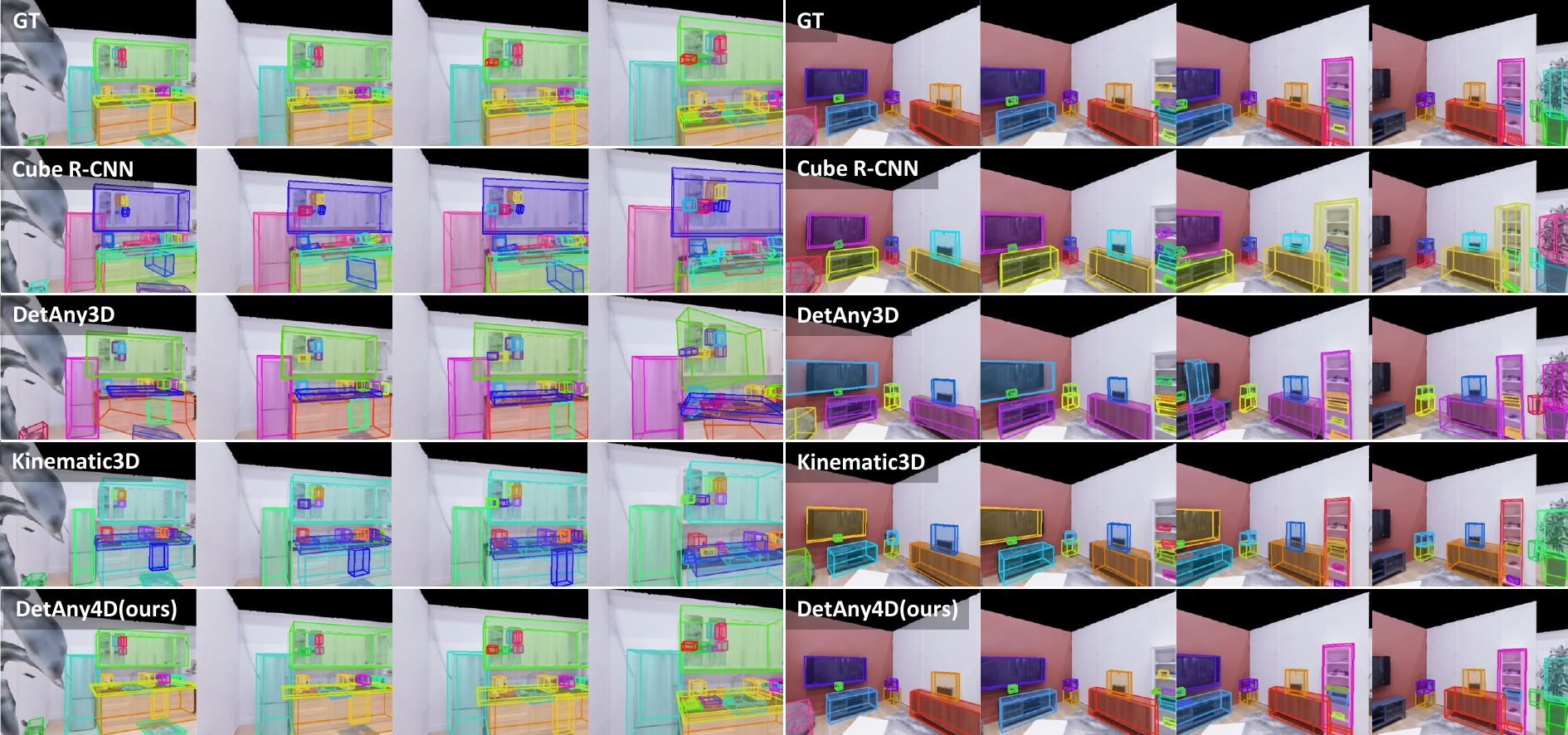}
\caption{More qualitative comparison results.}
\label{fig::exp6}
\end{figure*}

\end{document}